\def\year{2018}\relax
\documentclass[letterpaper]{article} %DO NOT CHANGE THIS
\usepackage{aaai18}  %Required
\usepackage{times}  %Required
\usepackage{algorithmicx,algpseudocode,algorithm} 
\newcommand*\Let[2]{\State #1 $\gets$ #2} 
\usepackage{times}  %Required 
\usepackage{amsmath}
\usepackage{amssymb} 
\usepackage{xcolor}
  
\newcommand{\add}{\textsf{addSet}} 
\newcommand{\low}{\textsf{low}}
\usepackage{helvet}  %Required
\usepackage{courier}  %Required
\usepackage{url}  %Required
\usepackage{graphicx}  %Required
\newcommand{\hide}[1]{} 
 
\newcommand{\focusOnMp}{\mbox{\textsf{focusOnMp}}}
\newcommand{\focusOnEc}{\mbox{\textsf{focusOnEc}}}
\newcommand{\focusOnOs}{\mbox{\textsf{focusOnOs}}} 
\newcommand{\focusOnImp}{\mbox{\textsf{focusOnImp}}} 
\newcommand{\focusOnLiq}{\mbox{\textsf{focusOnLiq}}}

\newcommand{\andC}{\textsf{and}}

\newcommand{\ryuta}[1]{{\color{magenta}{ryuta:#1}}} 
 
\newtheorem{theorem}{Theorem} 
\newtheorem{proposition}{Proposition} 
\newtheorem{example}{Example}

\begin{document}
\title{Abstractly Interpreting Argumentation Frameworks for  
Sharpening Extensions}
\author{Ryuta Arisaka and J{\'e}r{\'e}mie Dauphin\\ 
   ryutaarisaka@gmail.com, jeremie.dauphin@uni.lu
}
%\keywords{abstract argumentation; abstract interpretation; cycle}
\maketitle 
\begin{abstract} 
    Cycles of attacking arguments pose non-trivial issues in Dung style 
argumentation theory, apparent behavioural difference 
between odd and even length cycles being a notable one. 
While 
a few methods were proposed for treating them, to - in particular - enable selection of 
acceptable arguments  
in an odd-length cycle when Dung semantics could select none, 
so far 
the issues have been observed from a purely argument-graph-theoretic 
perspective. 
Per contra, we consider argument graphs together with a certain 
lattice like semantic structure over arguments e.g. ontology. 
As we show, the semantic-argumentgraphic hybrid theory  
allows us to apply {\it abstract interpretation}, 
a widely known methodology in static program analysis, to 
formal argumentation. With this, 
even where no arguments in a cycle could be selected sensibly, 
we could 
say more about arguments acceptability of an argument framework 
that contains it. In a certain sense, 
we can `verify' Dung extensions with respect to 
a semantic structure 
in this hybrid theory, to consolidate our confidence 
in their suitability. By defining the  theory, 
and by making comparisons to existing approaches, we ultimately 
discover that whether Dung semantics, or an alternative semantics such as cf2, is adequate or problematic 
depends not just on an argument graph but also on the semantic relation 
among the arguments in the graph.  

\end{abstract} 

\section{Introduction}              

Consider the following scenario: the members of a board of directors are gathered in a meeting to decide the future general strategy of their company.  
\begin{itemize} 
        \item $a_1$: ``We should focus on improving our business organization 
structure,  because it determines 
             our economic conduct." (\mbox{\textsf{focusOnOs}} 
             and \mbox{\textsf{OsDeterminesEc}}, to shorten) is 
             advanced by one member.  
          \item $a_2$: ``We should focus on improving our
             market performance, 
             because it determines 
             our business organization structure." 
             (\mbox{\textsf{focusOnMp}} and 
             \mbox{\textsf{MpDeterminesOs}}) is 
            then advanced by another member, as an attack on $a_1$.  
         \item $a_3$: ``We should focus on improving our
             economic conduct, 
             because 
             it determines our market performance." 
             (\mbox{\textsf{focusOnEc}} and 
             \mbox{\textsf{EcDeterminesMp}}), is then given 
             in response to $a_2$. 
         \item The first member attacks $a_3$, however, with $a_1$,
      to an inconclusive argumentation. 
          \item $a_5$: ``Our firm needs 1 billion dollars revenue 
       this fiscal year.", meanwhile, is an argument expressed  
           by another member.  
          \item $a_4$: ``Let our company just sink into bankruptcy!" 
             (\mbox{\textsf{focusOnLiq}}),
            another member impatiently declares in response, 
           against which, however,  
            all the first three speakers promptly express dissent with 
            their arguments. 
\end{itemize} 
\begin{center} 
    \includegraphics[scale=0.11]{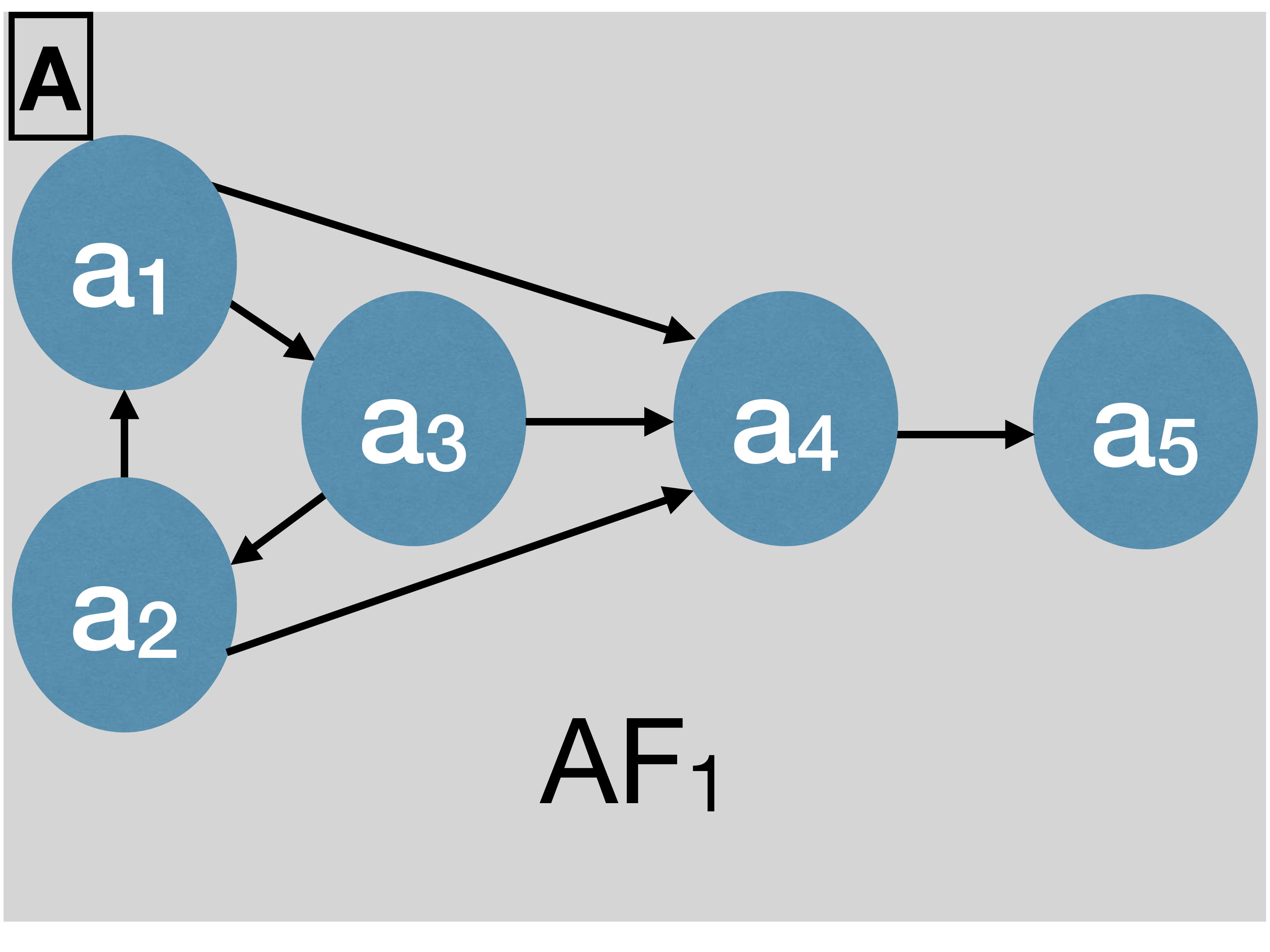}    
    \includegraphics[scale=0.11]{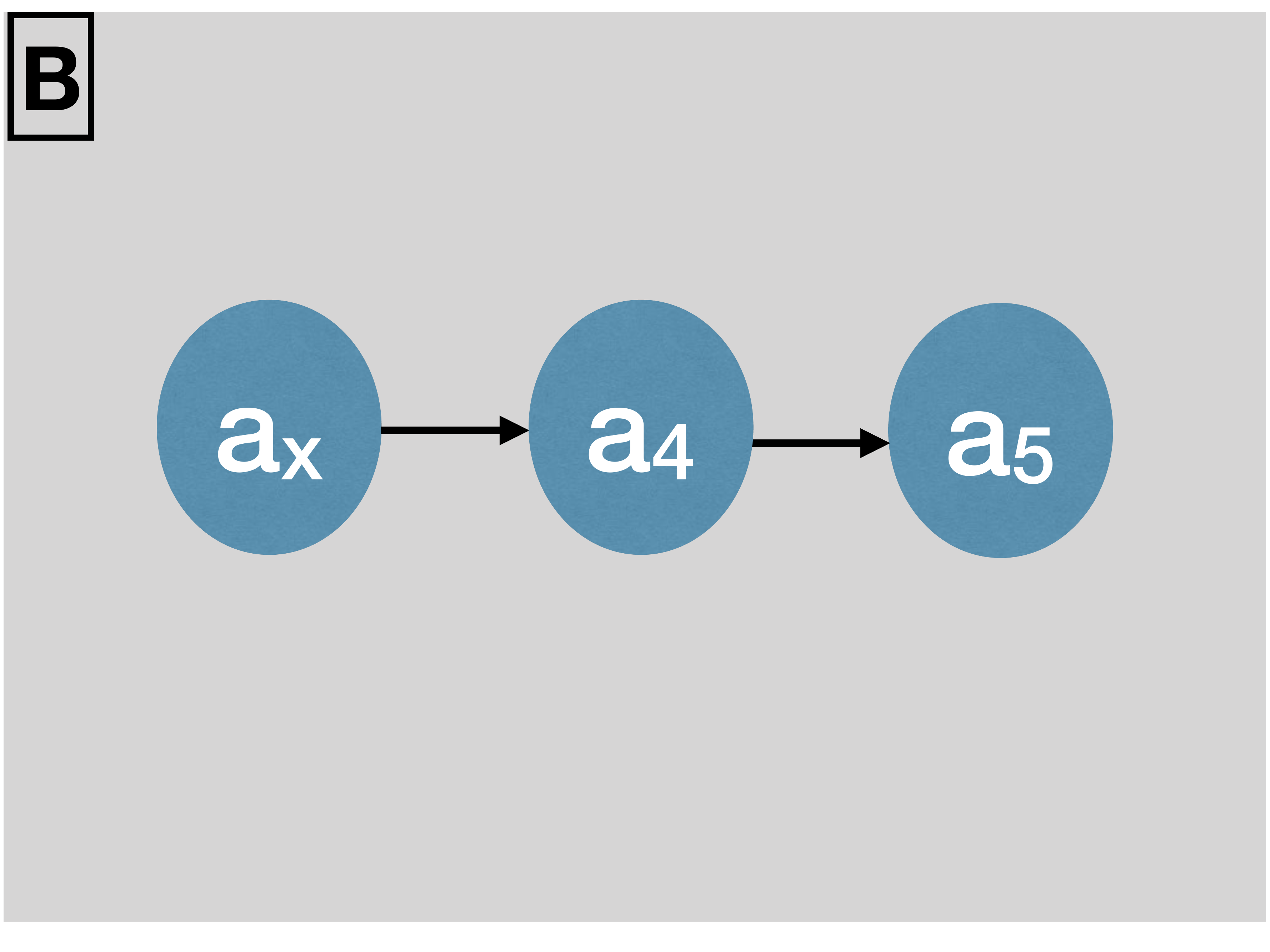} 
\end{center}   
\noindent We can represent this argumentation as $AF_1$ of Figure A.  
In Dung's abstract argumentation theory \cite{Dung95}, an admissible 
set of arguments is such that (1) no argument in the set is 
attacking an argument of the same set and (2) 
each argument attacking an argument in the set is attacked back
by some argument in the set. 
Clearly, there is no 
non-empty admissible set in $AF_1$. 
With labelling \cite{Caminada06} (an argument is \textsf{in} if all its attackers are \textsf{out}, is \textsf{out} if there exists an attacker 
that is \textsf{in}, and is \mbox{\textsf{undecided}}, otherwise), every 
argument is labelled \mbox{\textsf{undecided}}, 
and so we gain almost no information on the 
acceptability of the arguments. \\ 
%We can represent this argumentation as in Figure \fbox{B}.  
%In Dung theory \cite{Dung95}, an admissible 
%set of arguments is such that (1) no argument 
%in the set is attacking another one in the set and (2) 
%each argument attacking an argument in the set is attacked back
%by some argument in the set. Clearly, there is no non-empty 
%admissible set in $AF$ in Figure \ref{figureXY} \fbox{A}. %\jeremie{With the double attacks, there are actually three non-empty admissible sets now.}
%Now,  by applying the methods for determining acceptable arguments in the framework, we get that no arguments are acceptable. 
%Furthermore, by applying Caminada labelling, \cite{Caminada06} (i.e. an argument is \textsf{in} if all its attackers are \textsf{out}, is \textsf{out} if there exists an attacker that is \textsf{in}, and is \mbox{\textsf{undecided}}, otherwise), we get that all arguments are \mbox{\textsf{undecided}}, even though our intuition tells us $a_4$ should be out!  
%With labelling \cite{Caminada06} (i.e. an argument is \textsf{in} if 
%all its attackers are \textsf{out}, is \textsf{out} if there exists 
%an attacker that is \textsf{in}, and is \mbox{\textsf{undecided}}, 
%otherwise), we get that all arguments are \mbox{\textsf{undecided}}, 
%and so we gain almost no information on the acceptability 
%of the arguments. \\
\indent However, notice that $a_1$, $a_2$ and $a_3$ 
are arguments for the benefit 
of their company's growth. So, by aggregating  
them into a new argument $a_x$: %We should 
%focus on our company's further growth (\mbox{\textsf{focusOnImp}}), 
``We should 
focus on our company's further growth" (\mbox{\textsf{focusOnImp}}), 
and by thus deriving the framework as in Figure B,
we could obtain some more useful information on the 
acceptability of arguments, namely, that $a_x$ is \textsf{in},
$a_4$ is \textsf{out}, and $a_5$ is \textsf{in}. Hence 
we have sharpened acceptability statuses of $a_4$ and, 
in particular, $a_5$ of $AF_1$. \\
%\begin{center}
%    \includegraphics[scale=0.07]{secondExample.pdf} 
%\end{center}  
%\begin{center} 
%   \includegraphics[scale=0.11]{paradox}
%   \includegraphics[scale=0.11]{paradoxAbs}
%\end{center}
%\indent Sharpening of acceptability judgement is 
%not just in the direction of gaining information; 
%it can also work to reduce spurious options. 
%Consider $AF_2$ in Figure C. 
%By labelling, we would conclude either that $\{a_6, a_9\}$ are \textsf{in} 
%and $\{a_7, a_8\}$ are \textsf{out}, that 
%$\{a_7, a_8\}$ are \textsf{in} and $\{a_6, a_9\}$ are \textsf{out}, or else that 
%$\{a_6, a_7, a_8, a_9\}$ are all $\textsf{undecided}$. Let us consider 
%an instance of $AF_2$ with: 
%\begin{itemize}
%   \item $a_6$: ``It is time for Tony Blair to step down.'' 
%   \item $a_7$: ``It is time for Tony Blair to step down.'' (or ``It is
%     time for Bambi to step down.'' 
%     or whatever that becomes a paraphrase of $a_6$ 
%    in 
%    the context of this argumentation)  
%    \item $a_8$: ``The Conservative will be the governing party.'' 
%    \item $a_9$: ``The Conservative will be the governing party.'' 
%    (or ``The Tories will be in power.'' or whatever that
%     becomes a paraphrase of $a_8$ in the context of this argumentation) 
%\end{itemize} 
%In this case, notice that the even-loop of $a_6$ and $a_7$ 
%%is nothing but a self-attacking argument, which 
%we could abstract into the argumentation framework as in Figure D. 
%We conclude that none of the arguments in Figure D are 
%acceptable, eliminating the other two errorneous options. 
\subsection{Abstract interpretation for cycles}     
What we saw is effectively abstract interpretation \cite{Cousot77}, 
a powerful methodology known 
in static program analysis to map concrete space semantics to 
abstract space semantics 
and to do inferences in the latter space to say 
something about the former space. The abstract semantics 
is typically coarser than the concrete semantics; 
in our example, the detail of what exactly their company 
should focus on was abstracted away. In return, we were 
able to conclude that $a_x$ is \mbox{\textsf{in}} and, 
moreover, that 
$a_4$ is \mbox{\textsf{out}}. Compared to 
existing approaches to deal with cycles e.g. 
\cite{Baroni05}, which gives \mbox{\textsf{in}} state  
to $a_5$ by enforcing acceptance of either of $a_1$, $a_2$ and $a_3$ 
to reject $a_4$, this approach that we propose does not require one to
accept any of the arguments within the cycle, even provisionally, in order to
be able to reject $a_4$, and thus accept $a_5$. 
By abstracting away some or all of the cyclic arguments, we avoid 
having to accept any of them while rejecting others. \\
%Meanwhile, in the second example, we took advantage of semantic
%information to tell that the 2-loop functions as a 1-loop so as not 
%to accept any of the arguments, avoiding a paradoxical conclusion. \\
\indent In general, abstract interpretation is applicable 
to cycles of any length. There 
can be more than one way of interpreting an argumentation 
framework abstractly, however, and the key for obtaining 
a good outcome 
is to find properties sufficiently fine for abstraction. 
For the attacks among $a_1$, $a_2$, $a_3$ and $a_4$ in $AF_1$,  
we observe that, specifically: $a_1$'s 
\mbox{\textsf{focusOnOs}}, $a_3$'s 
\mbox{\textsf{focusOnEc}}, and $a_2$'s 
\mbox{\textsf{focusOnMp}} 
attacks $a_4$'s \mbox{\textsf{focusOnLiq}}; and 
$a_1$'s \mbox{\textsf{focusOnOs}} 
$a_3$'s \mbox{\textsf{focusOnEc}} and 
$a_2$'s \mbox{\textsf{focusOnEc}} form a cycle of attacks. 
For these,
the semantic information fine enough to abstract  
$a_1$, $a_2$, $a_3$ into $a_x$ is shown below (only 
\mbox{\textsf{focusOnX}} expressions are explicitly stated here).
%\begin{center}
%\includegraphics[scale=0.2]{space.pdf} 
%\end{center}  
\renewcommand\thefigure{\Alph{figure}}
\setcounter{figure}{2}
\begin{figure}[!h]
\begin{center}
\includegraphics[scale=0.13]{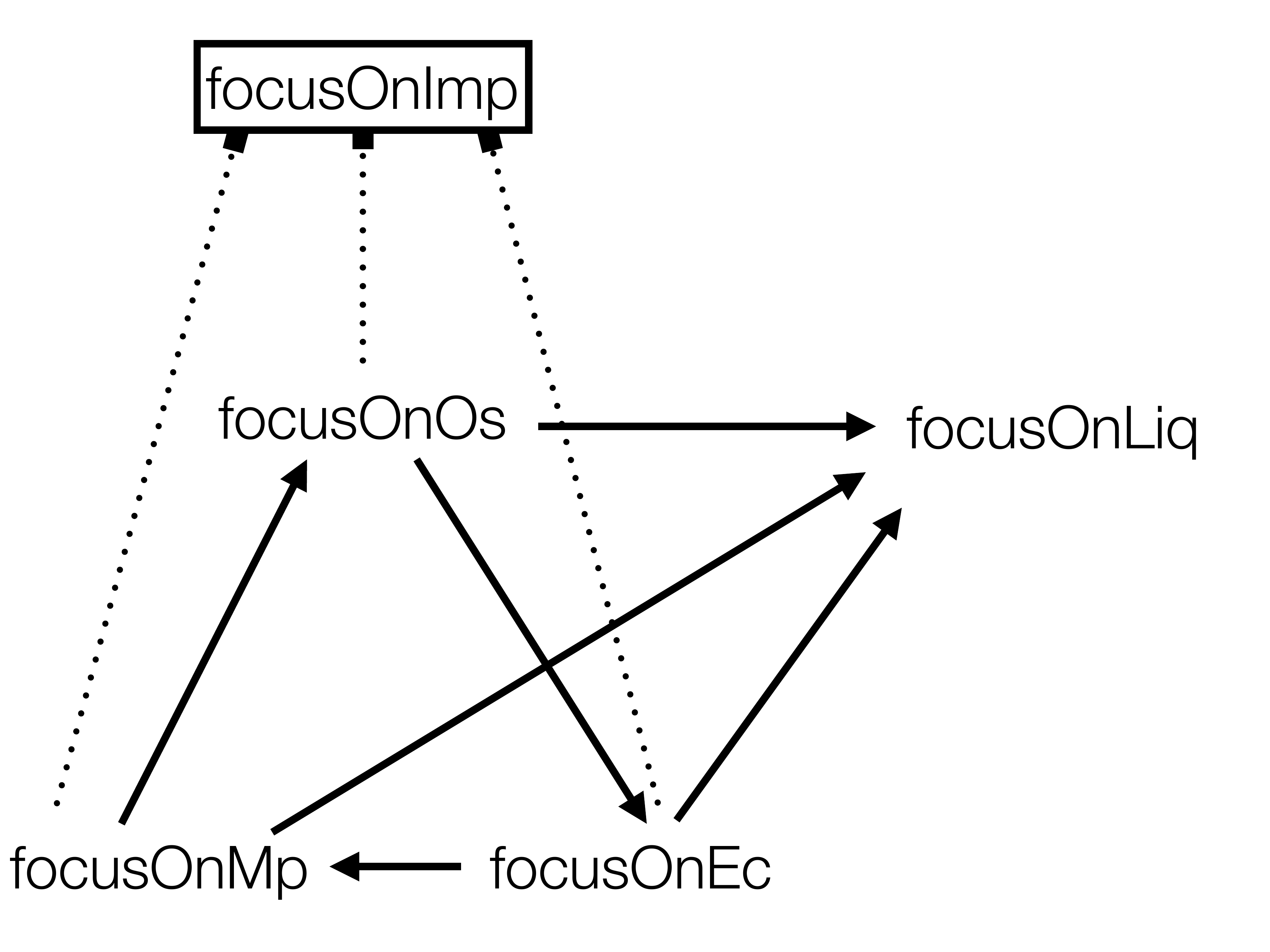}  
   \caption{\normalfont $AF$ and some ontological abstract-concrete relation 
     over its arguments. {\focusOnImp} is more abstract an argument than 
     \focusOnOs, \focusOnMp, and \focusOnEc. Neither 
     of the three is more abstract or more concrete 
     than the other two. {\focusOnLiq} is not a concrete  
     instance of \focusOnImp.} 
   \label{space} 
\end{center} 
\end{figure} 
In this figure, $a_x$ (\mbox{\textsf{focusOnImp}}: focus on further 
growth), sits as 
more abstract an argument of 
$a_1$, $a_2$ and $a_3$. 
While there may be other alternatives for 
abstracting them, \mbox{\textsf{focusOnImp}}  
belongs to a class of good abstractions, 
as it satisfies the property that 
the three arguments but not \mbox{\textsf{focusOnLiq}} 
fall into it, which gives justification 
as to why the attack relation in the initial argumentation framework 
is expected to preserve 
    in the abstractly interpreted argumentation 
    framework (from $a_x$ to $a_4$). This is 
    what we might describe the condition of {\it attack-preservation}. \\
\indent Further, 
      the three concrete arguments exhibit  
      a kind of competition for the objective:   
       their company's business focus. 
      While 
    ``organisation structure'', ``market performance'', and 
    ``economic conduct'' all vie for it and in that sense they 
      indeed oppose, neither of them 
     actually contradicts the objective, which is why abstraction 
     of the three arguments is possible here. \\
%  \ryuta{Put any extra here to emphasise the objectives 
%      we like to achieve here. For instance, how Dung semantics
%      may be `verified', how abstract and concrete space extensions 
%     relate.  Later.} \\
\indent We will formulate abstract interpretation for argumentation
     frameworks, the first study of the kind, as far as we are aware. 
    We will go through technical preliminaries (in Section 2), and 
    develop our formal frameworks and make comparisons 
     to Dung preferred and cf2 semantics (in Section 3), 
     before drawing conclusions. 
    The  discovery we ultimately make is that whether 
    Dung preferred or cf2 semantics is adequate or problematic  
    depends not only on the argumentation framework's structure, but also  
    on the semantic relation between its arguments. 
    We will show that our 
    methodology
    is one viable way 
    of enhancing accuracy in judgement as to 
    which set of arguments should be accepted.

\section{Technical preliminaries}  

    \subsection{Abstract argumentation} 
    Let $\mathcal{A}$ be a class of abstract entities.   
An \emph{argumentation framework} according to Dung's argumentation theory is a 2-tuple $(A, R)$ 
for $A \subseteq_{\text{fin}} \mathcal{A}$ and 
$R: A \times A$. Let $a$ with a subscript refer to a member 
of $A$, and let $A$ with or without a subscript refer to a subset 
of $A$. 
An argument $a_1$ is said to \emph{attack} another argument $a_2$ 
if and only if, or iff, $(a_1, a_2) \in R$. A subset $A_1$ 
is said to \emph{accept}, synonymously to \emph{defend}, $a_x$ 
iff, for each $a_y$ attacking $a_x$,
it is possible to find some $a_z\in A_1$ 
such that 
$a_z$ attacks $a_x$. A subset $A_1$ is said to 
be: \emph{conflict-free} iff no element of $A_1$ attacks an element 
of $A_1$; an \emph{admissible set} iff it is conflict-free and 
defends all the elements of $A_1$; 
and a \emph{preferred set (extension)} iff it is a set-theoretically maximal admissible set.
There are other classifications to admissibility,  and 
an interested reader will find details in \cite{Dung95}. 
An argument is \emph{skeptically accepted} 
iff it is in all preferred sets  
and \emph{credulously accepted} iff it is in at least one preferred set.
\subsection{Order and Galois connection for 
abstract interpretation}  
Let $L_1$ and $L_2$ (each) be an ordered set, 
ordered in $\sqsubseteq_1$ and $\sqsubseteq_2$ respectively.
Let $\alpha$ be an abstraction function that maps 
each element of $L_1$ onto an element of $L_2$, 
and let $\gamma$ be a concretisation function that 
maps each element of $L_2$ onto an element of $L_1$.  
$\alpha(l_1)$ for $l_1 \in L_1$ is said to be 
an abstraction of $l_1$ in $L_2$, 
and $\gamma(l_2)$ for $l_2 \in L_2$ is said to 
be a concretisation of $l_2$ in $L_1$. If 
$\alpha(l_1) \sqsubseteq_2 l_2$ implies $l_1 \sqsubseteq_1 \gamma(l_2)$ 
and vice versa for every $l_1 \in L_1$ and every $l_2 \in L_2$, then the pair of $\alpha$ and $\gamma$ 
is said to be a Galois connection. Galois connection 
is contractive: $\alpha \circ \gamma(l_2) \sqsubseteq_2 l_2$ 
for every $l_2 \in L_2$, and extensive: $l_1 \sqsubseteq_1 \gamma \circ 
\alpha(l_1)$ for every $l_1 \in L_1$. Also, both $\alpha$ and $\gamma$ 
are monotone with $\alpha \circ \gamma \circ \alpha = \alpha$
and $\gamma \circ \alpha \circ \gamma = \gamma$.  
An ordered set $L_1$, ordered by a partial order $\sqsubseteq_1$, is a complete lattice  
just when it is closed under join and meet for every $L_1' \subseteq L_1$. Every finite lattice 
is a complete lattice. 
%Let a set $L_3$ 
%be ordered in $\leq_3$ with an abstraction function $\alpha_2$  
%that maps each element of $L_2$ onto an element of $L_3$, 
%and a concretisation function $\gamma_2$ 
%that maps each element of $L_3$ onto an element of $L_2$. 
%If both $(\alpha, \gamma)$ and $(\alpha_2, \gamma_2)$ are
%a Galois connection, 
%then so is 
%$(\alpha_2 \circ \alpha, \gamma \circ \gamma_2)$. \\
\hide{ \indent Let $L$ be a set partially ordered by an order $\leq$. 
$L_1 (\subseteq L)$   
is said to be an ideal iff: (1) $L_1$ is non-empty; 
(2) if $x \in L_1$ and if $y \leq x$, then $y \in L_1$; 
and (3) if $x, y \in L_1$, then $x \leq z$ and $y \leq z$ 
for some $z \in L_1$. An ideal $L_1$ is said to be proper 
iff $L_1$ is not $L$ itself. \\}
%Let $L_1$ and $L_2$ be a complete lattice (see \cite{Davey02} 
%for properties of a lattice)
%with a
%partial order $\leq_1$ and $\leq_2$ respectively, then 
%$\bigtriangledown: L_2 
%\times L_2 \rightarrow L_2$ is said to be a widening iff  
%(1) $\gamma(x) \leq_1 \gamma(x \bigtriangledown y)$; (2)  
%$\gamma(y) \leq_1 \gamma(x \bigtriangledown y)$; 
%and (3) if $x_1 \leq_2 x_2 \leq_2 \cdots$ is an increasing chain, 
%then $x_1 \leq_2 {x_1 \bigtriangledown x_2} \leq_2 {(x_1 \bigtriangledown x_2) 
%    \bigtriangledown x_3} \leq_2 \cdots$ is a 
%not strictly increasing increasing chain.  

\section{Argumentation frameworks for abstraction}   

While, for our purpose, 
Dung's theory is not expressive enough, all 
we have to do is to detail  
the components of the tuple so that 
we gain access to some internal information 
of each argument. 
\subsection{Lattices}   
\begin{figure}[!h] 
\begin{center}
    \includegraphics[scale=0.177]{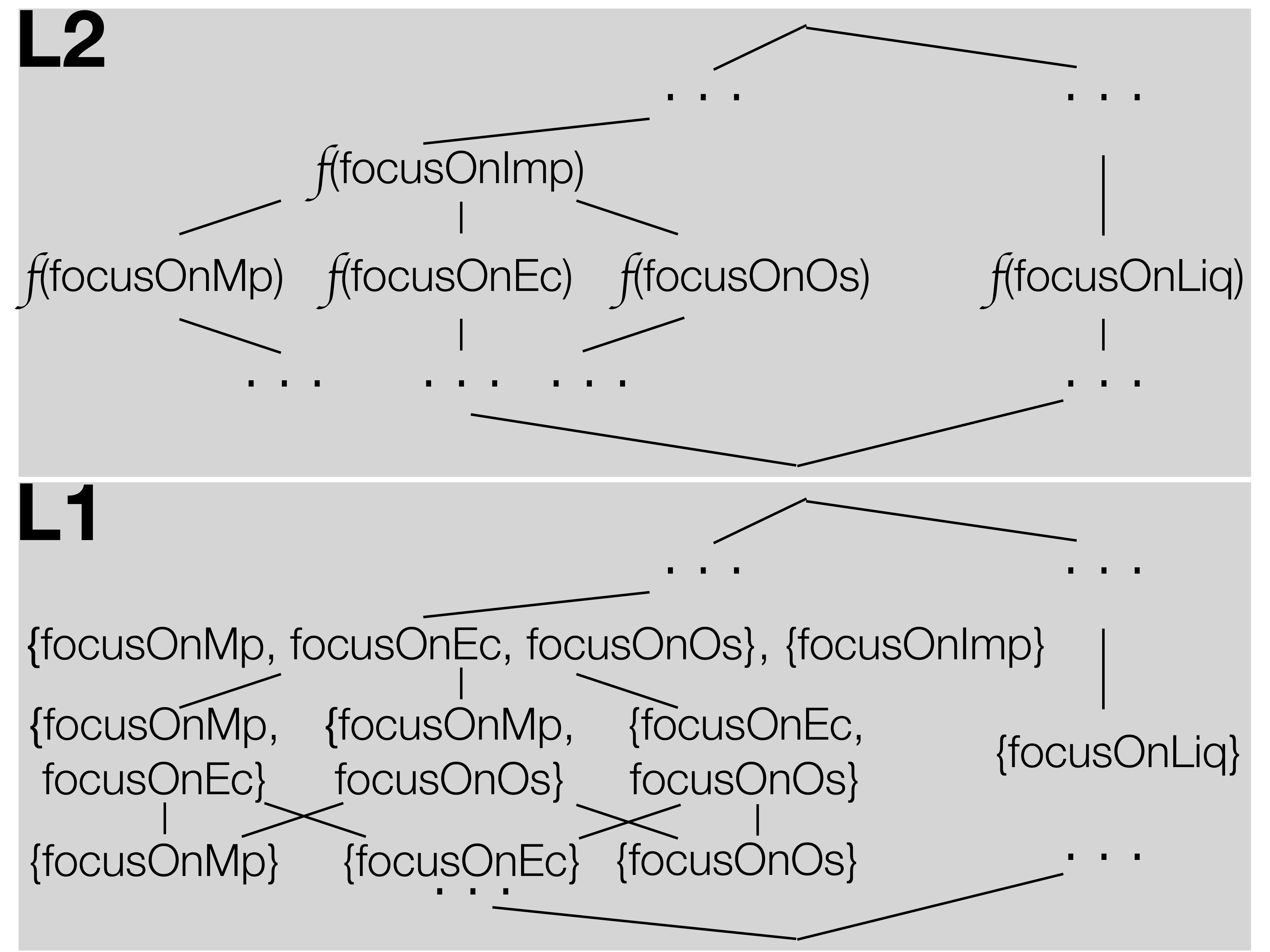} 
\end{center}  
\caption{\normalfont {\small Illustration of a concrete lattice and an abstract 
lattice.}}
\label{visualisation} 
\end{figure}

Let \mbox{$(L'_2, \sqsubseteq, \bigvee, \bigwedge)$} be a 
finite lattice. %\jeremie{You used $\leq_1$ and $\leq_2$ in the section just before for the orderings of $L_1$ and $L_2$, we should maybe stick to a common symbol throughout the sections for consistency.}
Let $\mathcal{E}$ be the class 
of expressions as abstract entities. 
We denote each element of $\mathcal{E}$ by $e$ with or without 
a subscript and a superscript. 
Those \mbox{$\textsf{focusOnMP}$} and others 
in our earlier example are expressions. 
Let \mbox{$f:    
\mathcal{E} 
\rightarrow L'_2$} 
be a function that maps an expression onto 
an element in the lattice. This function 
is basically a semantic interpretation 
of $\mathcal{E}$, which could be 
some chosen ontology representation with annotations 
of general-specific relation among entities. For example, 
in Introduction, \mbox{\textsf{focusOnImp}}  
was more general than \mbox{\textsf{focusOnMp}}, 
\mbox{\textsf{focusOnEc}} and \mbox{\textsf{focusOnOs}}, 
which should enforce 
\mbox{\textsf{focusOnImp}} mapped 
onto an upper part in $L'_2$ than 
the three, i.e. $f(\mbox{\textsf{focusOnEc}}), 
f(\mbox{\textsf{focusOnMp}}), f(\mbox{\textsf{focusOnOs}})\linebreak
\sqsubseteq f(\mbox{\textsf{focusOnImp}})$. In the rest, 
rather than $L'_2$ itself, we will 
talk of the sub-complete-lattice $L_2$ of all 
$f(e)$ for $e \in \mathcal{E}$ as well as its top and its bottom. \\
\indent $\mathcal{E}$ form abstract space arguments with 
the relation as defined in $L_2$. 
Concrete space arguments, in comparison, are just a 
set of expressions that can possibly be arguments. 
Let ${\low: L_2  \rightarrow 2^{L_2}}$ be such that:
$\low(l_2) := \{l_2\}$ if $l_2 = \bigwedge L_2$ (the bottom element); 
else
$\low(l_2) := \{x \in L_2 \mid  x \sqsubset l_2\ \andC\ \nexists y \in L_2 \text.x \sqsubset y \sqsubset l_2 \}$. 
%\jeremie{I rewrote this in a more formal way, hope this makes it clearer and more compact too. Also put $y \sqsubset l_2$, otherwise there always exists $y = l_2$.} 
%it is the set of all $x \in L_2$ satisfying 
%both (1) $x \sqsubset l_2$ 
%and (2) there exists no $y \in L_2$ such that $x \sqsubset y \sqsubseteq
%l_2$. 
We let \mbox{$(L_1, \subseteq', \bigcup, \bigcap)$} 
be another complete lattice 
where $L_1 := 2^{\mathcal{E}}$ and $\subseteq'$ satisfies: 
\begin{itemize} 
%\jeremie{Are you sure about this "all elements that match the description?" It seems to me that this would make $L_1$ simply contain all subsets of $\mathcal{E}$ and make the second requirement irrelevant? Also I was thinking we could maybe reformulate it as such:
%Let \mbox{$(L_1, \subseteq', \bigcup, \bigcap)$} be another complete lattice such that:
%\begin{itemize}
%	\item $L_1 \subseteq 2^\mathcal{E}$
%	\item if $l_1 \in L_1$ and $l'_1 \subseteq \mathcal{E}$ s.t. $\bigvee_{e' \in l'_1} f(e') \sqsubseteq f(e)$ for some $e \in l_1$, then $(l_1 \setminus \{e\} \cup l'_1 \in L_1)$
	\item $x \subseteq' y$ if $x \subseteq y$. 
	\item ${x \subseteq' y}$ and ${y \subseteq' x}$ 
	iff: ${x = \{e_1, \ldots, e_n\}}$ %for some $e_1, \ldots, e_n \in \mathcal{E}$
	and \\${y =} \{e_1, \ldots, e_{i-1}, e'_1, \ldots, e'_m, e_{i+1}, 
     \ldots, 
	e_n\}$ with \\ $\low(f(e_i)) = \{e'_1, \ldots, e'_m\}$. 
\end{itemize} 
%The first requirement looks a bit more formal and I think the second one is a bit clearer this way, and also it might be more compact to define the whole lattice in one go rather that separating the set and order, although that doesn't matter too much.}
%\jeremie{Why not just write $\{e_1, \ldots, e_n, e'_1, \ldots, e'_m\}$? Is $e_i$ not in the first set? If that's the case then I think writing $\{e_1, \ldots, e_{i-1}, e_{i+1}, \ldots, e_n, e'_1, \ldots, e'_m\}$ would make it clearer.} \ryuta{$e_i$ is replaced by $e'$s. I do not mind which way, 
%that is a matter of taste :D} 
 The lattices shown in Figure \ref{visualisation} 
illustrate the second condition. Notice that $\low(f(\mbox{\textsf{focusOnImp}})) = 
\{f(\mbox{\textsf{focusOnMp}}), 
     f(\mbox{\textsf{focusOnEc}}),
     f(\mbox{\textsf{focusOnOs}})\}$ in $L_2$.\linebreak
$\{\mbox{\textsf{focusOnImp}}\}$ and 
 {\small $\{\mbox{\textsf{focusOnMp}}, 
     \mbox{\textsf{focusOnEc}},
     \mbox{\textsf{focusOnOs}}\}$} are equivalent in  
    $L_1$ which is indeed a quotient lattice. 
This equivalence reflects the following interpretation of  
ours of 
expressions. Any expression  
$e_1$ has concrete instances $e_2, \ldots, e_i$ if  
$f(e_2), \ldots, f(e_i)$ are children of $f(e_1)$ in abstract lattice.  
If, here, $f(e_2), \ldots, f(e_i)$ 
are all the children of $f(e_1)$, our interpretation is that  
mentioning $f(e_1)$ is just a short-hand of mentioning  
all $f(e_2), \ldots, f(e_i)$, i.e. both mean the same thing with 
respect to the structure of $L_2$.  
It is because of this that we  place all equivalent sets of expressions 
at the same node in $L_1$. 
%\jeremie{Isn't it better to stick with the $focusOnX$ example here?}
\subsection{Argumentation frameworks}
We call an expression with an ID - which we 
    just take from $\mathcal{A}$ - 
    an argument-let, so the class of all argument-lets 
    is $\mathcal{A}^b: \mathcal{A} \times \mathcal{E}$.    
    Each argument-let shall be denoted by $a^b$ with 
    a subscript.  \\ 
     \indent   We update Dung's $(A, R)$ 
    into $(A^b, R^b)$ 
    for $A^b \subseteq_{\text{fin}} \mathcal{A}^b$, 
and $R^b:  
    A^b \times A^b$. To readers interested in knowing compatibility 
    with  
    Dung's argumentation frameworks,  
    Dung's argument corresponds to 
    a set of all argument-lets in $A^b$ 
    that share the same ID. For example,  
    we may have ${A^b = \{(a_1, e_1), (a_1, e_2), 
            (a_1, e_3), (a_2, e_1), (a_2, e_4)\}}$, in which case 
    $A^b$ maps into $A = \{a_1, a_2\}$ if projected 
    into Dung's 
    argumentation framework. 
    For compatibility of attack relation, too, 
    Dung's $(a_1, a_2) \in R$, i.e. $a_1$ attacks $a_2$, (assuming 
    that both $a_1$ and $a_2$ are in $A$) 
    corresponds to  
    $((a_1, e_1), (a_2, e_2)) \in R^b$ for 
    some $e_1$ and some $e_2$ (assuming 
    both $(a_1, e_1)$ and $(a_2, e_2)$ 
    are in $A^b$).  
    For convenience hereafter, by argument $a$ with or without a subscript,  
    we refer to a set of argument-lets in $A^b$ that 
    share   
    the ID $a$. We do not consider any more structured arguments
    than a finite subset of $\mathcal{A}^b$ in this work; further 
    structuring, while of interest, 
    is not the main focus, which 
    is left to a future work. 
    All notations 
    around extensions: conflict-freeness, acceptance and defence, 
    admissible and preferred sets, are carried over here for 
    arguments (note, not for argument-lets). 
    \subsection{Abstraction and concretisation}    
Now, let $\alpha: L_1 \rightarrow L_2$ 
    be the abstraction function, 
    and let $\gamma: L_2 \rightarrow L_1$ 
    be the concretisation function.  
    We require: $\alpha(l_1) = \bigvee_{e_u \in l_1} f(e_u)$; 
    and $\gamma(l_2) = \{x \in \mathcal{E} \ | \ f(x) \in \low(l_2)\}$. 
    Intuition is as we described earlier in 3.1.  
    Note $\gamma(l_2)$ is an empty set when     
    $\low(l_2)$ does not contain any $f(e)$ for $e \in \mathcal{E}$. 
    We say that $e_x$ 
    is the best abstraction of $\{e_1, \ldots, e_n\}$ 
    iff $f(e_x) = \alpha(\{e_1, \ldots, e_n\})$, 
    but more generally we say that
    $e_x$ is an abstraction of $e_1, \ldots, e_n$  iff  
    $\alpha(\{e_1, \ldots, e_n\}) \sqsubseteq f(e_x)$. 
    We say that $\{e_1, \ldots, e_n\}$ is the most 
    general
    concretisation of $e_x$ iff $\{e_1, \ldots, e_n\} = 
    \gamma(f(e_x))$. More generally, 
    we say that $\{e_1, \ldots, e_n\}$ is a concretisation of $e_x$ 
    iff $\{e_1, \ldots, e_n\} \subseteq' \gamma(f(e_x))$.    
      \begin{proposition}[Galois connection] \normalfont 
        For every $l_1 \in L_1$ 
        and every $l_2 \in L_2$,  
        we have $\alpha(l_1) \sqsubseteq l_2$ 
        iff $l_1 \subseteq' \gamma(l_2)$. 
    \end{proposition}   
    \textbf{Proof} 
    \textsf{If}: Suppose $l_2 \sqsubset \alpha(l_1)$, i.e. 
    $l_2 \sqsubset \bigvee_{e_u \in l_1} f(e_u)$ by definition of $\alpha$.  $\sqsubset$ is a standard abbreviation. 
    Then we have $\gamma(l_2) \subset' l_1$, contradiction. 
    Suppose $l_2$ and $\alpha(l_1)$ are not comparable 
    in $\sqsubseteq$, then clearly $l_1 \not\subseteq' \gamma(l_2)$, 
    contradiction. 
    \textsf{Only if}:  Suppose $\gamma(l_2) \subset' l_1$, 
    then there exists at least one $e$ in $l_1$  
    which is not in any set equivalent 
    to $\gamma(l_2)$ under $\subseteq'$. Then by definition of $\alpha$, 
    we have $l_2 \sqsubset \alpha(l_1)$, contradiction.  \hfill$\Box$ \\
    \begin{example} \normalfont 
    In Figure \ref{visualisation}, 
     $\low(f(\focusOnImp)) =     
     \{\focusOnMp, \focusOnEc, \focusOnOs\}$. 
    We see that, for instance,  
    $\{\mbox{\textsf{focusOnMp}}, \mbox{\textsf{focusOnEc}}\}$ is 
     mapped to $f(\mbox{\textsf{focusOnImp}})$ by $\alpha$ 
    as  $\alpha(\{\mbox{\textsf{focusOnMp}}, \mbox{\textsf{focusOnEc}}\}) =   {f(\mbox{\textsf{focusOnMp}}) \vee f(\mbox{\textsf{focusOnEc}}})$. 
    $\mbox{\textsf{focusOnImp}}$ is hence (the best) abstraction 
     of $\{\mbox{\textsf{focusOnMp}}, \mbox{\textsf{focusOnEc}}\}$.  
 Meanwhile,  $\gamma(f(\mbox{\textsf{focusOnImp}})) = 
      \{\textsf{focusOnMp}, \textsf{focusOnEc},$\linebreak
  $\textsf{focusOnOs}\} = X$. 
       Since $(\alpha, \gamma)$ is a Galois connection,   
       $\alpha(X) = f(\focusOnImp)$ again. 
    \end{example} 
    \indent Let $E \subseteq \mathcal{E}$ with or without a subscript  
    denote a set of expressions. %\jeremie{What is the difference between $E$ and $L_1$? We have introduced some good notation for $L_1$ already such as $\alpha(l_1)$, $\subseteq'$, $\cup$ and $\cap$, so why re-use them? Especially $\alpha$ could be useful since you are using $\bigvee_{e \in E_{a_x}}f(e)$ a lot later on.} 
    Each argument-let comes with a singleton set of expression, 
    so an argument comes with a set of expressions.  
When we write $E_{a_x}$, we mean to refer to  
all expressions associated with $a_x$. 
    For abstraction, 
    we say that an argument $a_x$ 
    is:  
    \begin{description}
        \item[abstraction-covering] 
    for a set of arguments $a_1, \ldots, a_n$   
    iff, if $e_x \in E_{a_x}$  
    is an abstraction of $E \subseteq 
    \{E_{a_1}, \ldots, E_{a_n}\}$,  
    then it is an abstraction 
    of $\bigcup_{1 \leq i \leq n} E'_{a_i}$ 
    where $E'_{a_i}$ is a non-empty subset of $E_{a_i}$. 
\item[abstraction-disjoint]  
    for a set of arguments $a_1, \ldots, a_n$  
    iff, for each $a_i$, $1 \leq i \leq n$,    
    if $e_k \in E_{a_x}$ is an abstraction of $e_u \in E_{a_i}$, 
    then $e_j \in E_{a_x}$, $j \not= k$, is not $e_u$'s abstraction.  
\item[abstraction-sound]  
    for a set of arguments $a_1, \ldots, a_n$ 
    iff,  
    for each $a_i$, $1 \leq i \leq n$, 
    there is no $e \in E_{a_i}$  
    that is not abstracted by  
    any $e \in E_{a_x}$. 
\item[abstraction-complete] 
    for a set of arguments $a_1, \ldots, a_n$ 
    iff, 
    for each $e \in a_x$, 
    $e$ is an abstraction of $E \subseteq 
    \bigcup_{1 \leq m \leq n} E_{a_m}$. 
\end{description}  
\begin{figure}[!h] 
        \begin{center} 
            \includegraphics[scale=0.115]{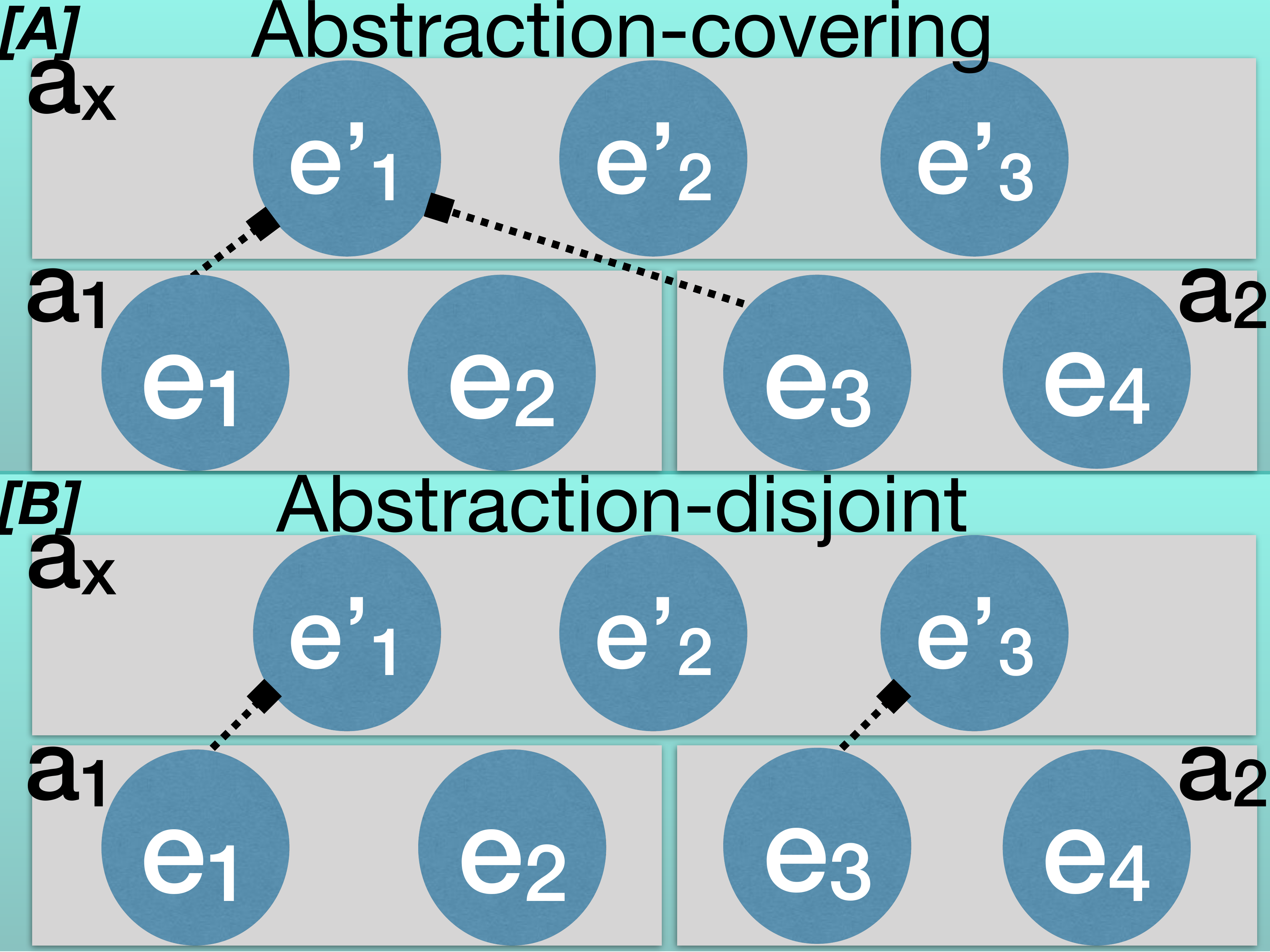} 
            \includegraphics[scale=0.115]{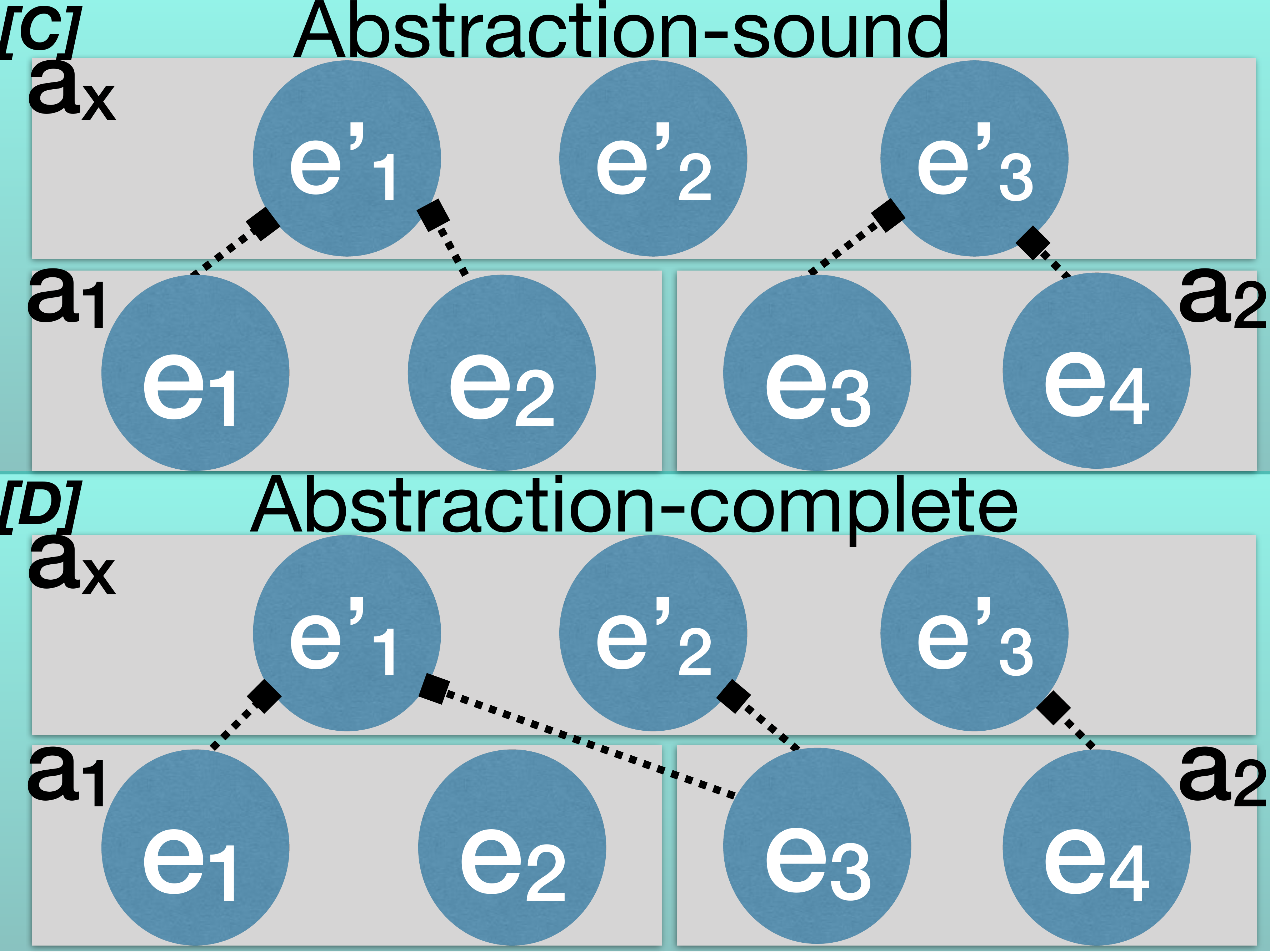}  
        \end{center}   
        \caption{\normalfont Illustration for 
            abstraction-covering-ness [A], abstraction-disjointness [B],
            abstraction-soundness [C] and 
            abstraction-completeness [D].}  
        \label{abstraction-figures} 
    \end{figure} 
Figure \ref{abstraction-figures} illustrates 
            these 4 conditions.  
            $a_x$ comes with ${\{e'_1, e'_2, e'_3\}}$, 
            $a_1$ with ${\{e_1, e_2\}}$, 
            and $a_2$ with ${\{e_3, e_4\}}$. 
            In [A], $a_x$ 
        is abstraction-covering for $\{a_1, a_2\}$ because 
        $e'_1 \in E_{a_x}$ abstracts from a non-empty subset 
        of both $E_{a_1}$ and $E_{a_2}$. If 
        $e'_1$ should abstract only from a non-empty subset 
        of $E_{a_1}$ but not of $E_{a_2}$, abstraction-covering-ness would not be satisfied. 
        In [B], $a_x$ satisfies abstraction-disjointness because
         any expression is abstracted at most by 
         one expression in $E_{a_x}$. If, here, $e'_1$ 
         should abstract both from $e_1$ and $e_2$, 
         this condition would fail to hold. 
                  In [C], $a_x$ satisfies abstraction-soundness 
because  
         all expressions in $E_{a_1} \cup E_{a_2}$ are abstracted. 
         If there should be even one expression  
         in $E_{a_1} \cup E_{a_2}$ that is not abstracted,
        this condition would fail to hold.  
In [D], 
$a_x$ satisfies abstraction-completeness because
         each expression in $E_{a_x}$ abstracts 
         expressions in $E_{a_1} \cup E_{a_2}$.  
         If there should be even one expression  
        in $E_{a_x}$ that does not abstract any
        expressions in $E_{a_1} \cup E_{a_2}$, this condition
would fail to hold. 
    \begin{proposition}[Independence] \normalfont  {\ }\\
        Let $\omega, \beta$ be 
        one of the propositions $\{$\mbox{$a_x$ is abstraction-covering}, 
        \mbox{$a_x$ is abstraction-sound},\\
        \mbox{$a_x$ is abstraction-disjoint}, 
        \mbox{$a_x$ is abstraction-complete}$\}$.   \\
    Then $\omega$ materially implies $\beta$ 
       iff $\omega = \beta$. 
    \end{proposition}    
    As for motivation of the four conditions,  
    our goal for abstraction of a given set $A$ of 
    arguments dictates that, in whatever manner  
    we may abstract, eventually we abstract 
    from all expressions of all the arguments in $A$.
    Hence 
    we have abstraction-soundness. However, consider  
    an extreme case where each $e'_i$ abstracts from 
    a single argument $a_1$. Then, certainly, 
    such abstraction weakens each member of $A$, but  
    there is no guarantee that the weakening 
    is a weakening of $A$, because  
    abstraction of $\{e_1, \ldots, e_n\}$ is 
    necessarily abstraction of each of $e_1, \ldots, e_n$, but 
    abstraction of $e_1$ is not necessarily that of $\{e_1, \ldots, e_n\}$. 
    Not abstracting from each and every $a_1 \in A$ 
    is problematic for this reason. 
    Abstraction-covering-ness is therefore desired.  
    Abstraction-disjointness discourages  
    redundancy in abstraction. 
    Abstraction-completeness ensures relevance 
    of abstraction to a given set of arguments to be abstracted. 
    In the rest, {\it whenever} we state 
    $a_u$ is an abstraction of  
    $\{a_1, \ldots, a_n\}$, $a_u$ will be assumed to be 
      abstraction-covering, abstraction-disjoint, 
     abstraction-sound and abstraction-complete 
     for them. We say that the abstraction is the best abstraction 
     iff it is the best abstraction of 
     all expressions associated with $\{a_1, \ldots, a_n\}$. 
     \begin{proposition}
         \normalfont 
        If  $a_x$ is the best abstraction 
        of  a set $A$ of arguments, then 
        every abstraction 
        of $A$ is an abstraction of $a_x$. 
     \end{proposition} 
      
     \begin{proposition}[Existence]\label{existence}  \normalfont 
         There exists at least one abstraction 
       for every set of arguments. 
    \end{proposition}   
    \textbf{Proof} 
        $L_2$ is a complete lattice.  \hfill$\Box$ \\\\
        \noindent However, 
        some abstraction, including the top element of 
        $L_2$ if it is in $\{f(e) \in L_2 \ | \ e \in \mathcal{E}\}$, 
        can be so general 
        that all arguments are abstracted by it. 
        In the first example in Introduction,  
        ``Argumentation is taking place." 
        could be such an argument, in which 
        one may not 
        be normally interested for reasoning about argumentation: 
        the whole point of argumentation theory is for us to be 
        able to judge which set(s) of arguments may be acceptable
        {\it when the others are unacceptable}, 
        so we should not trivialise a given argumentation by 
        a big summary argument.
        \subsubsection{Conditions 
            for conservative abstraction} 
\begin{figure}[!h]   
\begin{center}  
          \includegraphics[scale=0.1]{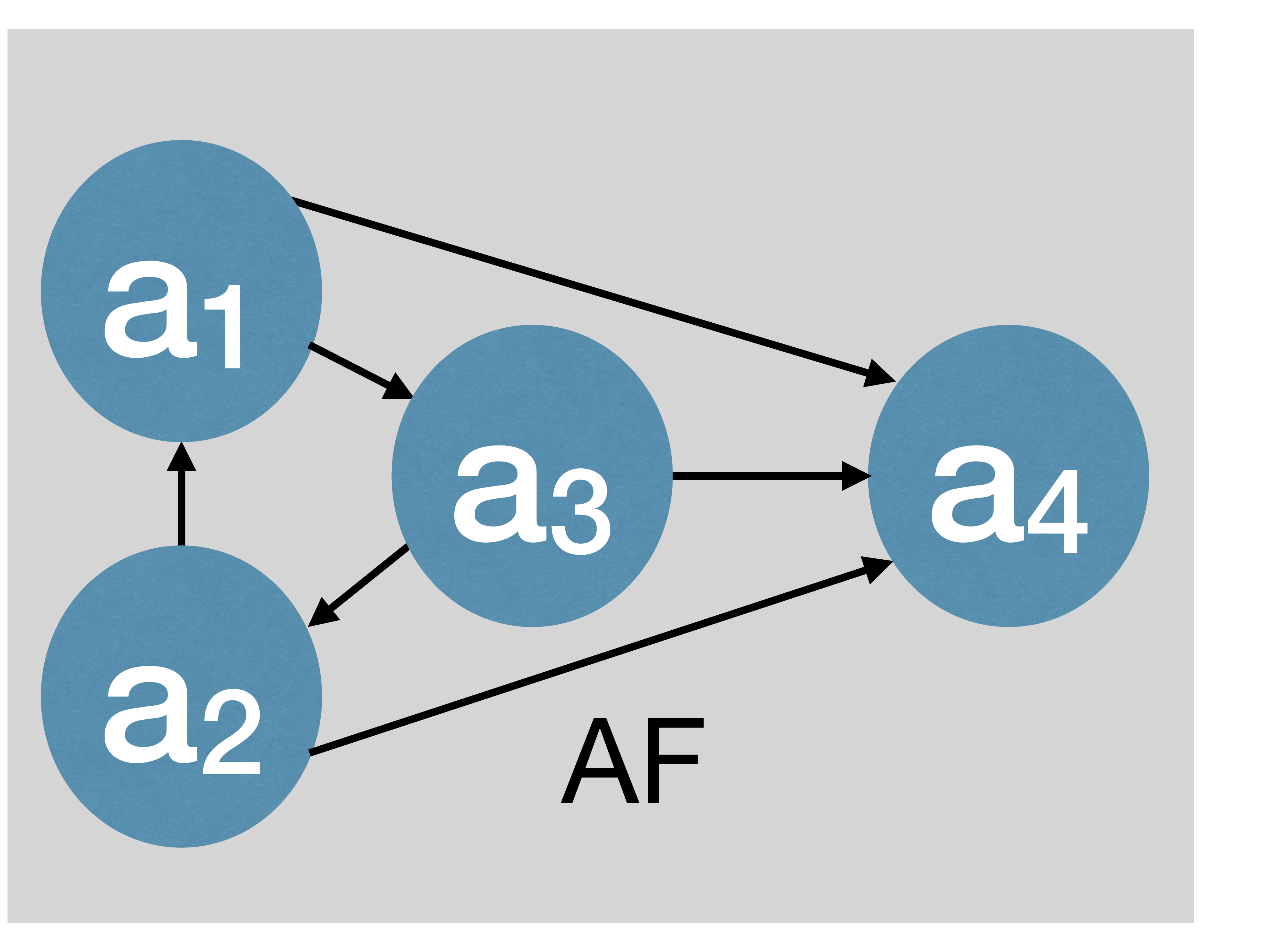} 
          \includegraphics[scale=0.1]{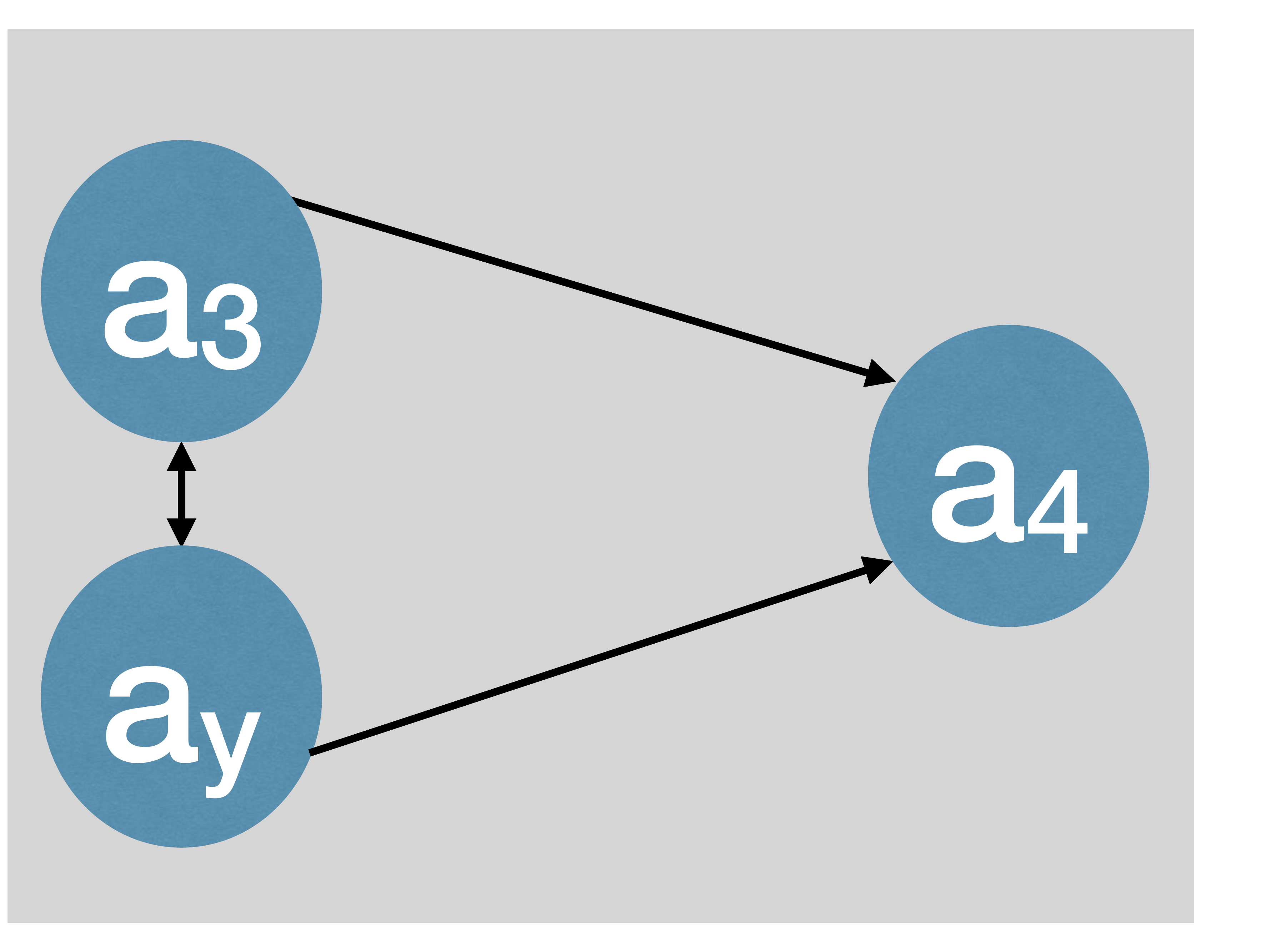}  
\end{center}     
          \caption{\normalfont Left: an argumentation framework 
          before abstraction. 
              Right: an argumentation framework after a compatible 
but  invalid abstraction. } 
          \label{intermediate} 
       \end{figure}

\begin{figure}[!h]
        \begin{center} 
           \includegraphics[scale=0.11]{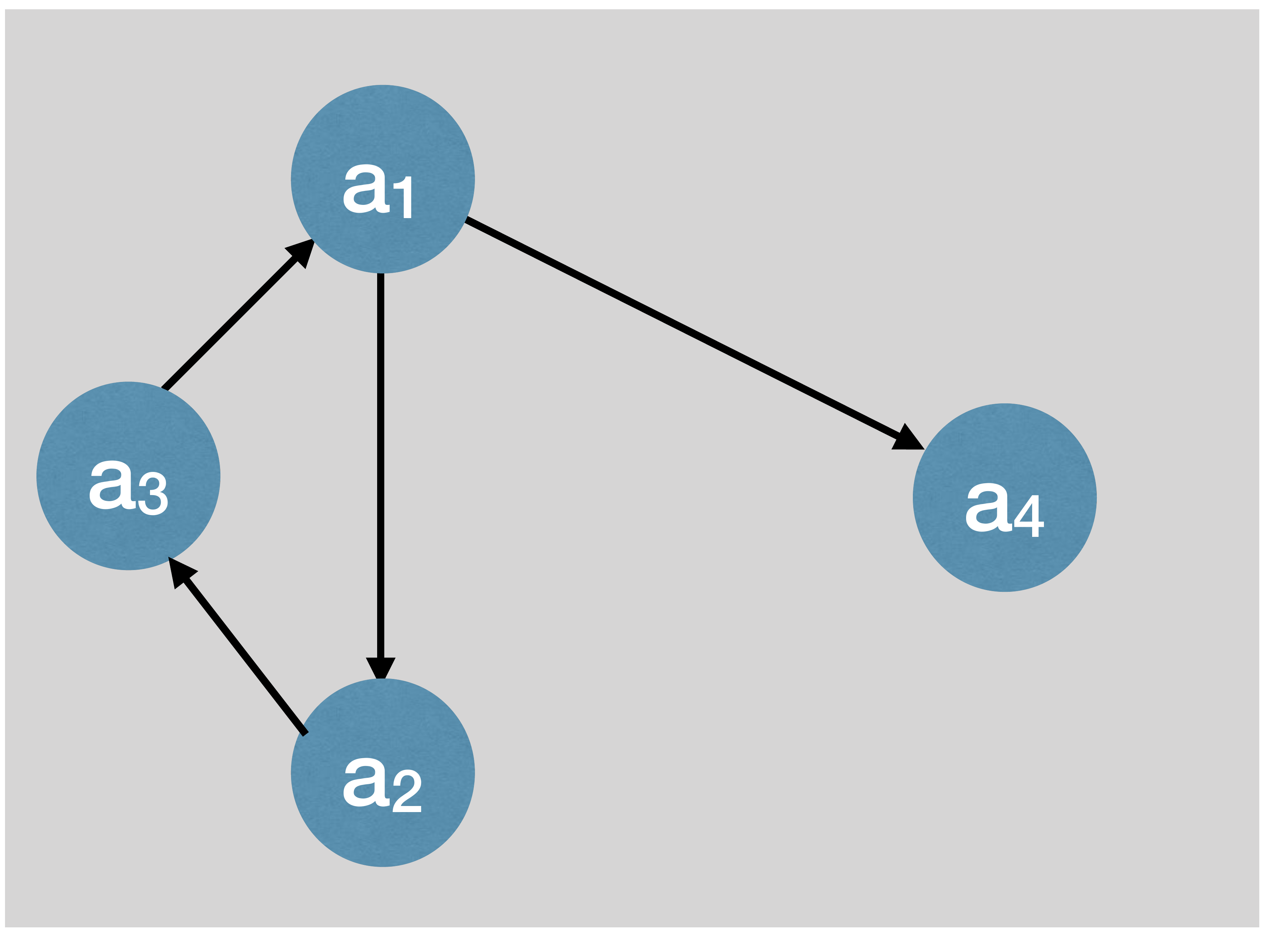} 
          \includegraphics[scale=0.11]{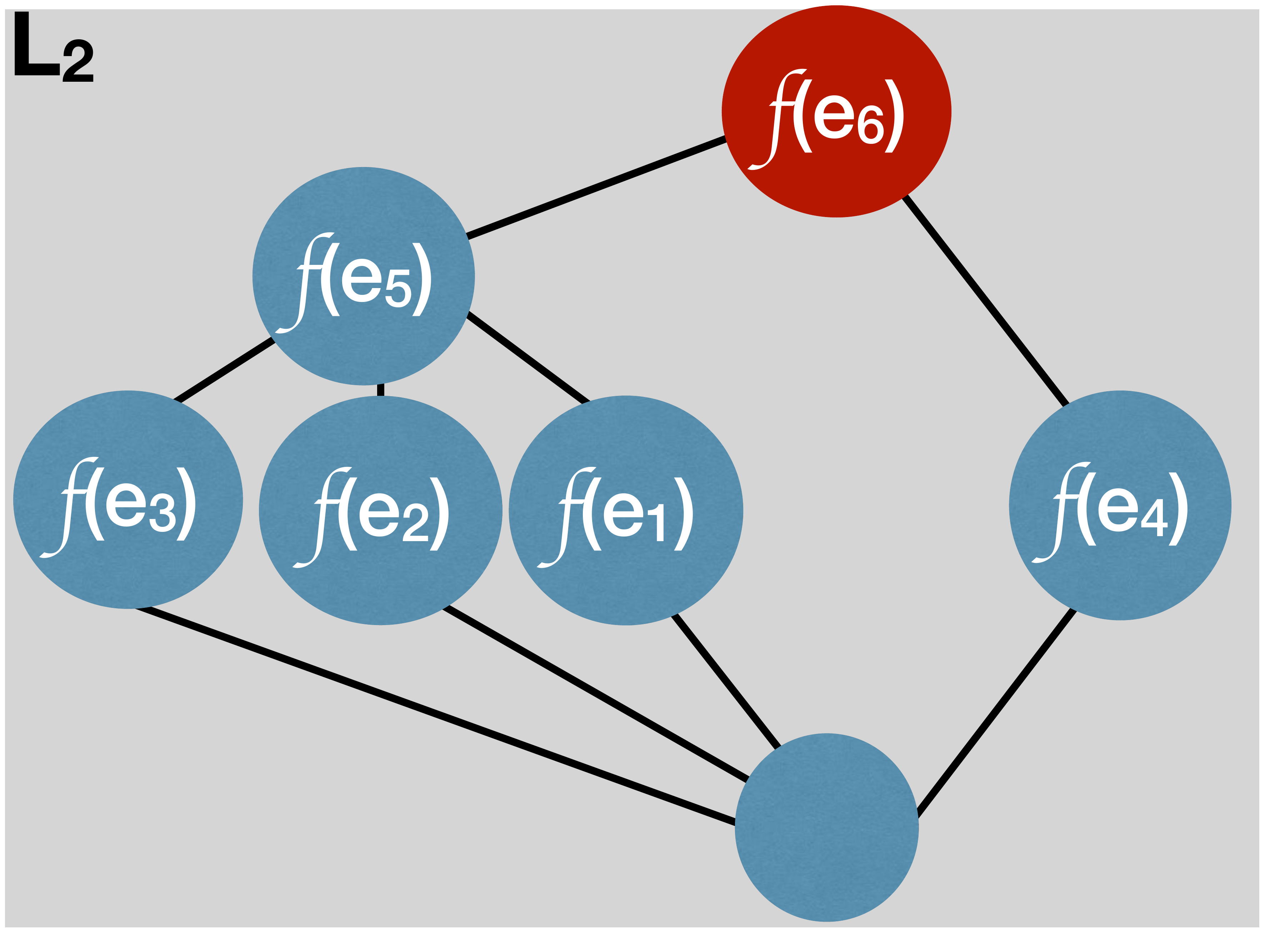} 
        \end{center}  
\caption{\normalfont Left: an argumentation framework with 4 arguments. 
$E_{a_i}$ is assumed to be
a singleton $\{e_i\}$. Right: 
abstract lattice $L_2$. $\alpha(a_i)$ is assumed to be $f(e_i)$.} 
\label{acillustrate} 
\end{figure}

        Hence, a few conditions ought to be defined 
        in order to ensure 
        conservative abstraction. 
        Assume an argumentation framework 
        $(A^b, R^b)$. We assume  
        that those elements of $L_2$ that are so abstract 
        that they could abstract all argument-lets 
        in $A^b$ into a single argument-let 
        are forming a non-empty upper set $M$ of $L_2$: 
        $M$ ($\subseteq L_2$) is an upper set iff, if $x \in M$ and 
        $x \sqsubseteq y$, then $y \in M$. 
        Intuition is that once we find some $f(e)$ in $L_2$ 
        that is so general, then any $f(e_1)$ 
        such that $f(e) \sqsubseteq f(e_1)$ is also. 
 For example, if  $f(\focusOnImp)$ in $L_2$ 
        is so general, then  
$f($``Argumentation is taking place."$)$ is also so general. \\
        \indent Let us say that 
        there is a path from an argument $a_1$ to 
        an argument $a_2$ iff either $a_1$ attacks $a_2$, 
        or else there is a path from $a_1$ 
        to some argument $a_3$ which attacks $a_2$. 
        Let us say that 
        a set $A_1$ of arguments is strongly 
        connected component in $(A^b, R^b)$ 
        iff (1) there is a path from any $a_1 \in A_1$ 
        to any $a_2 \in A_1$ and (2) there exists 
        no $A_1 \subset A_x \subset A^b$ such that 
        $A_x$ satisfies (1).  
        For an argumentation framework $(A^b, R^b)$, we define that 
        abstraction $a_x$ of a set $A_1$ of arguments 
        is: {\it valid} iff there exists   
        a strongly connected component $A_s \subseteq A^b$ 
        such that: (1) $A_1 \subseteq A_s$; and 
        (2) there exists no $A_1 \subset A_2 \subset A_s$ 
       such that $a_x$ is an abstraction of $A_2$ 
      (abstraction is over as many members of a strongly
connected component as possible);  
        {\it non-trivial} iff   
        $\alpha(E_{a_x}) \not\in M$ 
(abstraction 
        cannot be too general); and 
        {\it compatible} iff: there exist
        no argument-lets $(a_1, e_1), (a_2, e_2) \in A$ that satisfy
        both (1) $((a_1, e_1), (a_2, e_2)) \in R^b$ and 
        (2) $f(e_1)$ and $f(e_2)$ are comparable in $\sqsubseteq$ 
        (abstraction cannot be over arguments that contain an attack
        from more abstract an argument on more concrete an argument or vice 
       versa).  \\
%     \begin{itemize}
%       \item (A) if   
%        there exist argument-lets $(a_1, e_1), (a_2, e_2) \in A$ 
%        such that $((a_1, e_1), (a_2, e_2)) \in R^b$, then either: 
%         \begin{itemize}   
%           \item  (A1) $f(e_1)$ and $f(e_2)$ are not comparable 
%        in $\sqsubseteq$ $\andC$ $a_x$ is not self-attacking
%            \item  or else (A2) 
%        $f(e_2) \sqsubseteq f(e_1)$ $\andC$ $a_x$ is self-attacking,  
%         \end{itemize} 
%       \item or else (B) $a_x$ is not self-attacking 
%    \end{itemize}    
%        (if abstraction 
%        is over arguments that contain an attack  
%        from a more abstract expression of an argument on a more concrete
%        expression of an argument, then the 
%        abstract argument is self-attacking,
%        or if abstraction is over arguments that contain 
%        no attacks in abstract-concrete relation, 
%        then the abstract argument is not self-attacking). \\\\
  \indent      What compatibility expresses is: 
       a pair of arguments $a_1$ and $a_2$ with an attack
       between them is not suited for abstraction if $a_1$ (or $a_2$)
        is more, if not equally, abstract than $a_2$ (or $a_1$).
       For justification, suppose firstly that $a_2$ is $a_1$.
      Then it is a self-attack. Let us suppose that 
     abstraction of $a_1$ and $a_2$ is feasible, then we can   
     get rid of all self-attacks by means of abstraction.
     But that would render all such self-attacks not playing
     any role in argumentation, which cannot be the case \cite{Baumann17}.  
     In a more general setting where $a_2$ is not $a_1$, if it is 
     $a_1$ that attacks $a_2$, given that $a_2$ is more concrete an argument
     of $a_1$, the attack is again a type of self-attack. Still, 
     it is not safe to compile away the attack by taking abstraction 
     of $a_1$ and $a_2$, because the abstraction is more, if not equally,
     abstract than $a_1$ which $a_2$ was attacking. The second condition of 
     validity is motivated in a way by compatibility.\footnote{We, however,
     have a more recent result on abstraction of self-attacks. 
     An interested reader can contact either of the authors for detail.} 
Let us consider 
        the example in Introduction again (a part of it is re-listed in Figure  
        \ref{intermediate} on the left). 
        There are three arguments in the cycle, and   
        $\alpha(E_{a_1})$, 
        $\alpha(E_{a_2})$  and 
        $\alpha(E_{a_3})$ are not comparable 
        in $\sqsubseteq$ (Cf. Figure \ref{visualisation}). %\jeremie{Here in particular, the only difference between the 3 is a sub-sub-subscript, it would be easier to read if we used $\alpha(E_{a_1})$, $\alpha(E_{a_2})$ and $\alpha(E_{a_3})$}
        The least upper bound of any two among the three, 
        by the way, 
        is the same element in $L_2$. 
        Hence, by taking abstraction of only two among them, 
        say $a_1$ and $a_2$,  
        we obtain an abstract space argumentation framework 
        as in the right figure of Figure \ref{intermediate}.  
       As the attacks between $a_y$ and $a_3$ 
        are both of abstract-concrete and of concrete-abstract, 
   the compatibility condition prevents any further 
      abstraction on this argumentation framework. This, however,
       is amiss, because such abstract-concrete (concrete-abstract) relation 
       among the participants of the cycle  
       were not present (they were not comparable in $\sqsubseteq$)
       in the original argumentation framework. 
       The validity condition precludes this anomaly. 
        \begin{proposition}[Independence]  \normalfont 
            Let $\omega, \beta$ 
            be one of the propositions: 
            $\{$$a_x$ is valid, $a_x$ is non-trivial, 
                $a_x$ is compatible$\}$.
            $\omega$ materially implies $\beta$ iff $\omega = \beta$.     
        \end{proposition}
        These are conditions that apply for 
        abstraction of a given set of arguments alone.   
        In an argumentation framework, however, 
        we also consider attacks 
        between a set of arguments and the arguments 
        that are not in the set. We say that 
        abstraction $a_x$ of a set $A_1$ of arguments 
        is {\it attack-preserving} iff,  
        for each $a^b \in A^b \backslash A_1$ and each 
        $a_1^b \in A_1$, if at least either
        $(a^b, a_1^b) \in R^b$ or 
        $(a_1^b, a^b) \in R^b$, then  
        $\alpha(E_{a_x})$ and $\alpha(E_{a_1^b})$ are not 
        comparable in $\sqsubseteq$ 
        (abstraction of  
        $a_x$ and external attackers/attackees shall not be
        in abstract-concrete (concrete-abstract) relation). 
        For intuition
        behind this condition, let us consider Figure \ref{acillustrate}. 
      For simplicity, let us assume that 
$E_{a_i}$, $1 \leq i \leq 4$, is a singleton  $\{e_i\}$. The abstract 
lattice $L_2$ is shown in Figure \ref{acillustrate}. 
See to it that the attack of $a_1$ on $a_4$ is not 
of absolute-concrete (concrete-absolute). Now, with (the best) abstraction 
of $a_1$, $a_2$ and $a_3$, we obtain $a_5$ with $E_{a_5} = \{e_5\}$.  
While, in general, there is no continuity between 
some argument $a_1$ attacking 
some argument $a_4$ and some abstraction $a_x$ of 
$a_1$ attacking $a_4$, an exception is taken when   
there exists some pivotal point in $L_2$ that strongly 
distinguishes $\alpha(E_{a_x})$ and 
$\alpha(E_{a_4}) = f(e_4)$ (which, by the definition 
of abstraction,  
means $a_1$ and $a_4$ are equally distinguished), $a_x$ to one 
group, and $a_4$ to another distinct group. In such a case, 
as the attack of $a_1$ on $a_4$ can be viewed as 
the attack of the group that $a_1$ belongs to on the group that 
$a_4$ belongs to, and as $a_1$ and $a_x$ belong to the same group, 
abstraction of $a_1$ into $a_x$ does not modify the attack. 
This strong distinction holds just when $\alpha(E_{a_x})$ and 
$\alpha(E_{a_4})$ are not comparable in $\sqsubseteq$. This justifies 
retention of the attack by $a_5$ on $a_4$ (the pivot is  
$f(e_6)$) in the abstract space
argumentation framework. \\
\indent        We say that abstraction $a_x$
        of a set $A_1$ of arguments is 
        {\it conservative} iff 
        it is valid, non-trivial, compatible, and 
        attack-preserving.  
\begin{theorem}\label{theorem1} \normalfont 
    For a given $(A^b, R^b)$, 
 if an abstraction $a_x$ of $A_1 \subseteq A^b$ is conservative, 
      then  each abstraction $a_y$ of $A_1$ such that 
      $a_x$ is an abstraction of $a_y$ is conservative. 
\end{theorem}  
\textbf{Proof} 
Suffice it to check the four conditions 
one by one. \hfill$\Box$ 
%\textbf{Proof.} Suffice it to check the four conditions 
%one by one. \hfill$\Box$ 
    \subsection{Computation of abstract space 
        argumentation frameworks from a concrete 
    space argumentation framework}       
\begin{algorithm}[!h]  
       {\small 
            \caption{\normalfont Computation 
                of the set of abstract 
                 space argumentation frameworks  
                 for a given concrete space argumentation 
                 framework}  
             \label{algorithm}  
             \begin{algorithmic}[1] 
                 \Require{$X$ is an argumentation framework, 
                   $X.\add(Y)$ adds 
                  the elements of $Y$ into $X$, but 
                  is assumed to 
                  discard duplicates.}  
                 \Statex 
                 \Function{DeriveAbs}{$X$}    
                 \Let{$\Sigma$} an empty set. 
                 \State{\Comment{abs.space.arg.framwrks to 
                         be added to $\Sigma$}}       
                 \State{$\Sigma.\add(X)$}  
                 \Comment{Initially only $X$ is in $\Sigma$} 
                 \Let{$\Gamma$} all distinct sets  
                 of arguments in $X$ that 
                 are strongly connected.     
                \ForAll{$A$ in $\Gamma$}         
                \Let{$\Sigma_1$} $\Sigma$  \Comment{Copy $\Sigma$} 
                \Let{$\Sigma$} an empty set. \Comment{Reset} 
                 \Let{$\Pi$} the set of 
                 all maximal subsets of $A$ 
                 that satisfy conservative abstraction.  
                 \While{$\Sigma_1$ is not empty}   
                 \While{$\Pi$ is not empty}    
                  \Let{$X_1$} the 1st element of $\Sigma_1$
                 \Let{$A_1$} the 1st element of $\Pi$ 
                 \Let{$a_x$} the best abstraction 
                 of $A_1$  
\State{Replace $A_1$ 
                   in $X_1$ with $a_x$, while 
                   preserving attacks.}    
                 \State{$\Sigma.\add(X_1)$}  
                 \State{Remove the 1st element of 
                     $\Pi$} 
                 \EndWhile 
                 \State{Remove the 1st element of $\Sigma_1$} 
                 \EndWhile
                 \EndFor 
             \State{\Return $\Sigma$}
             \EndFunction 
             \end{algorithmic}  
         }
      \end{algorithm}  
      \noindent All abstract space argumentation frameworks  
         with conservative and best abstraction 
         can be computed for a given argumentation 
         framework, $\mathcal{E}$, $f$, $L_2$ and  
          $M \subseteq L_2$ 
         with Algorithm \ref{algorithm} which, informally, just keeps    
         replacing, where possible at all, a part of, or an entire, 
         cycle with an abstract argument for all possibilities. 
 Concerning Line 9, for a set of arguments $A_1$ in a given argumentation 
         framework, we say that $A_2 \subseteq A_1$ 
         is a maximal subset of $A_1$ that satisfies  
         conservative abstraction iff (1) 
         the best abstraction of $A_2$ is conservative 
         and (2) 
         there exists 
         no $A_3$ that satisfy both (2A): $A_2 \subset A_3 \subset A_1$ 
         and (2B): the best abstraction of $A_3$  
         is conservative.   
   \begin{proposition}[Complexity] \normalfont 
          Algorithm 1 runs at worst in exponential time. 
     \end{proposition}   
     \textbf{Proof.} 
       Strongly connected components are known to be 
       computable in linear time (Line 5).   
       Line 9 is computable at worst in exponential time. 
       With $n$ argument-lets (with possibly less than $n$ 
       arguments), we can over-estimate that
       the for loop executes at most 
       $n$ times, the 1st while loop at most  
       $(_nC_{\lceil n/2 \rceil})^n$ times, and the 2nd while loop at most 
       $_nC_{\lceil n/2 \rceil}$ times.  \hfill$\Box$ 
        \subsection{Preferred sets in concrete and abstract spaces}         
We now subject preferred sets in concrete space  
to
those in abstract space  
for more clues on 
arguments acceptability in 
concrete space. 
     Let us denote Algorithm \ref{algorithm} by $g_{\alpha}$, and 
   a function with [inputs = a set of argumentation frameworks] 
    and [output = a set of all preferred sets for 
     each given argumentation framework] 
   by $g_p$ (the procedure 
   can be found in the literature). 
   Further, let $g_{\gamma}$ be a projection function  
   with [inputs =  a set of sets of sets of arguments (i.e.  
all preferred sets for each argumentation framework)] and 
[outputs = a set of sets of sets of arguments], 
   with the following description. 
   Let $\sigma$ be a function with [inputs = a set of 
    sets of arguments $\times$ arguments]  
and [outputs = a set of sets of arguments] such that 
$\sigma(X, A) = \{\{A_1\} \ | \ \exists A_2 \in X.A_1 = A_2 \cap A\}$. 
Then $g_{\gamma}^{(A^b, R^b)}(Z) = \{\{X_1\} \ | \ \exists X \in Z.
   X_1 = \sigma(X, A^b)\}$. For example,   
   if $A^b = \{a_1, a_2\}$, then 
    $g^{(A^b, R^b)}_{\gamma}(\{\{\{a_1, a_3\}\},\{\{a_4\},\{a_2, a_5\}\}\})$
    is $\{\{\{a_1\}\}, \{\{a_2\}\}\}$. 
\begin{figure}[!h]  
      \begin{center}
          \includegraphics[scale=0.15]{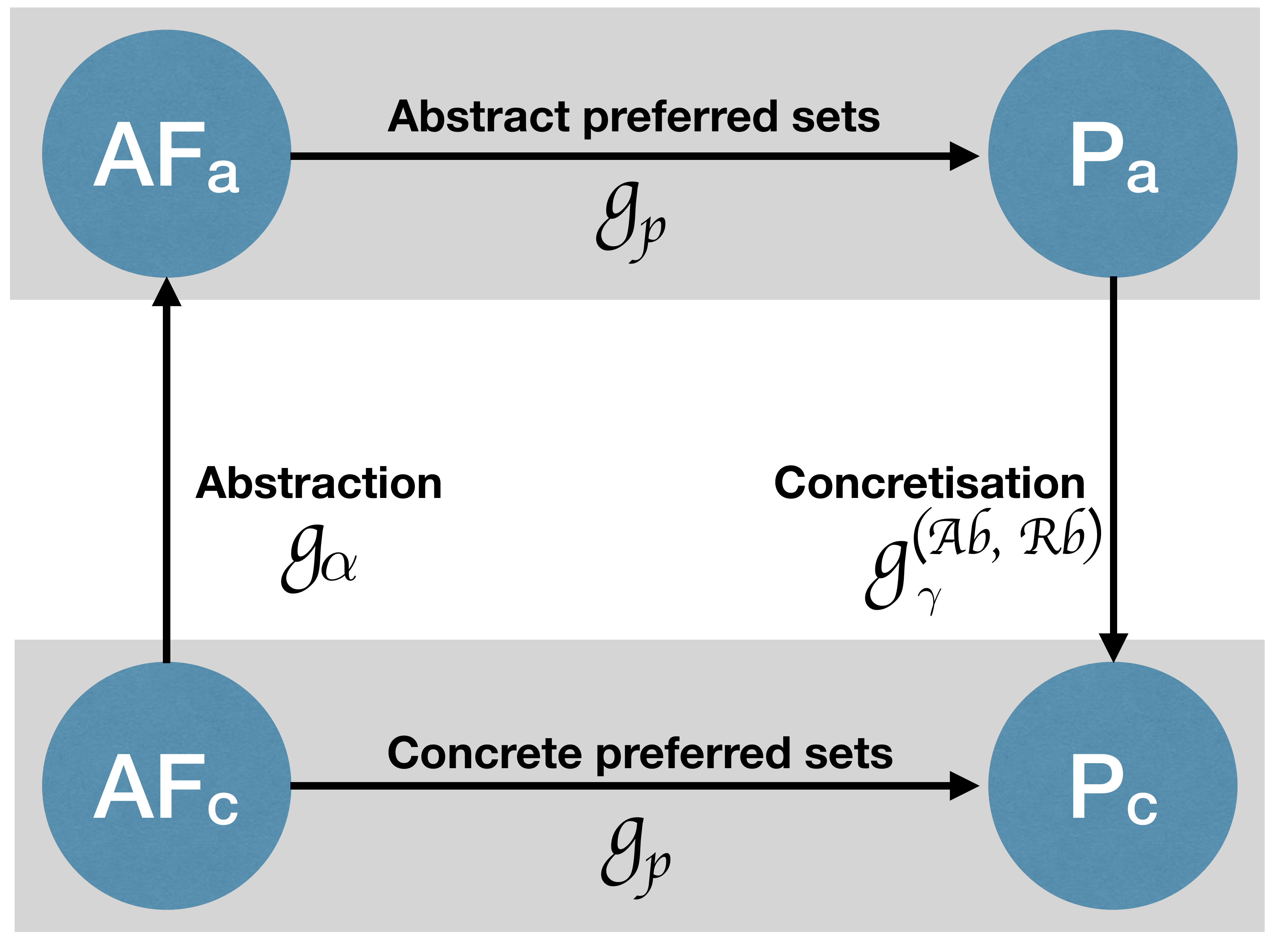} 
      \end{center} 
      \caption{\normalfont Relating abstract and concrete preferred 
          extensions. It is assumed that $AF_c = (A^b, R^b)$. }  
      \label{transformer} 
  \end{figure}    
  Figure \ref{transformer} illustrates on one hand 
    $g_{p}(\{AF_c\})$ for an argumentation 
   framework $AF_c$ in concrete space, which 
   gives us all preferred sets of $AF_c$, 
   and on the other hand 
   $g_{\gamma}^{(A^b, R^b)} \circ g_{p} 
   \circ g_{\alpha}(\{AF_c\})$, which 
   also gives us a set of all preferred sets  
  in concrete space but 
   through abstraction. 
   The abstract transformations proceed by 
   transforming the given concrete space 
argumentation framework into a set of abstract space 
argumentation frameworks ($g_{\alpha}(\{AF_c\})$), 
deriving preferred sets for them ($g_p \circ g_{\alpha}(
\{AF_c\})$), 
and projecting them to concrete space preferred sets 
($g_{\gamma}^{(A^b, R^b)} \circ g_p \circ g_{\alpha}(\{AF_c\})$) so that comparisons to 
the preferred sets obtained directly within concrete space can 
be done. In particular, we can learn: (1) an argument 
deemed credulously/skeptically acceptable within 
concrete space is {\it positively approved} by abstract space preferred sets, 
thus we gain more confidence in the set members 
being acceptable; (2) arguments not deemed 
acceptable within
concrete space, i.e. those that are not in any preferred set, 
are {\it negatively approved} also by abstract space preferred sets,  
thus we gain more confidence in those arguments not acceptable. 
But also: (3) arguments deemed credulously/skeptically 
acceptable within concrete space 
may be {\it questioned} when their acceptability is not inferred  
from any abstract space preferred set; 
and, on the other hand, (4) arguments deemed not acceptable within concrete space 
may be credulously/skeptically {\it implied} by abstract space preferred set(s). 
To summarise formally, given an argumentation 
framework $AF:(A^b, R^b)$, we say that an argument that is deemed credulously/skeptically 
acceptable in concrete space is: 
     \begin{description}
         \item[+approved] iff, for some/every element $X$ 
             of $g_{\gamma}^{(A^b, R^b)} \circ g_{p} \circ g_{\alpha}(\{AF_c\})$, 
            it belongs to some/every element $A$ of $X$. 
	 \item[questioned] iff, for every element $X$ 
         of $g_{\gamma}^{(A^b, R^b)}
          \circ g_{p} \circ g_{\alpha}(\{AF_c\})$, 
            it belongs to no element $A$ of $X$.  
     \end{description} 
    And we say that an argument that is deemed not acceptable 
    in concrete space is: 
    \begin{description}
       \item[-approved] iff, for every 
         element $X$ of 
         $g^{(A^b, R^b)}_{\gamma} \circ g_p \circ g_{\alpha}(\{AF_c\})$, 
          it belongs to no element $A$ of $X$. 
       \item[credulously/skeptically implied] iff, 
             for some/every element $X$ of 
            $g_{\gamma}^{(A^b, R^b)} \circ g_p 
            \circ g_{\alpha}(\{AF_c\})$, it belongs 
      to some/all member(s) $A$ of $X$. 
    \end{description}    
\subsection{Comparisons to Dung preferred semantics and 
    cf2 semantics, and observations}  
\begin{figure}
    \begin{center} 
      \includegraphics[scale=0.11]{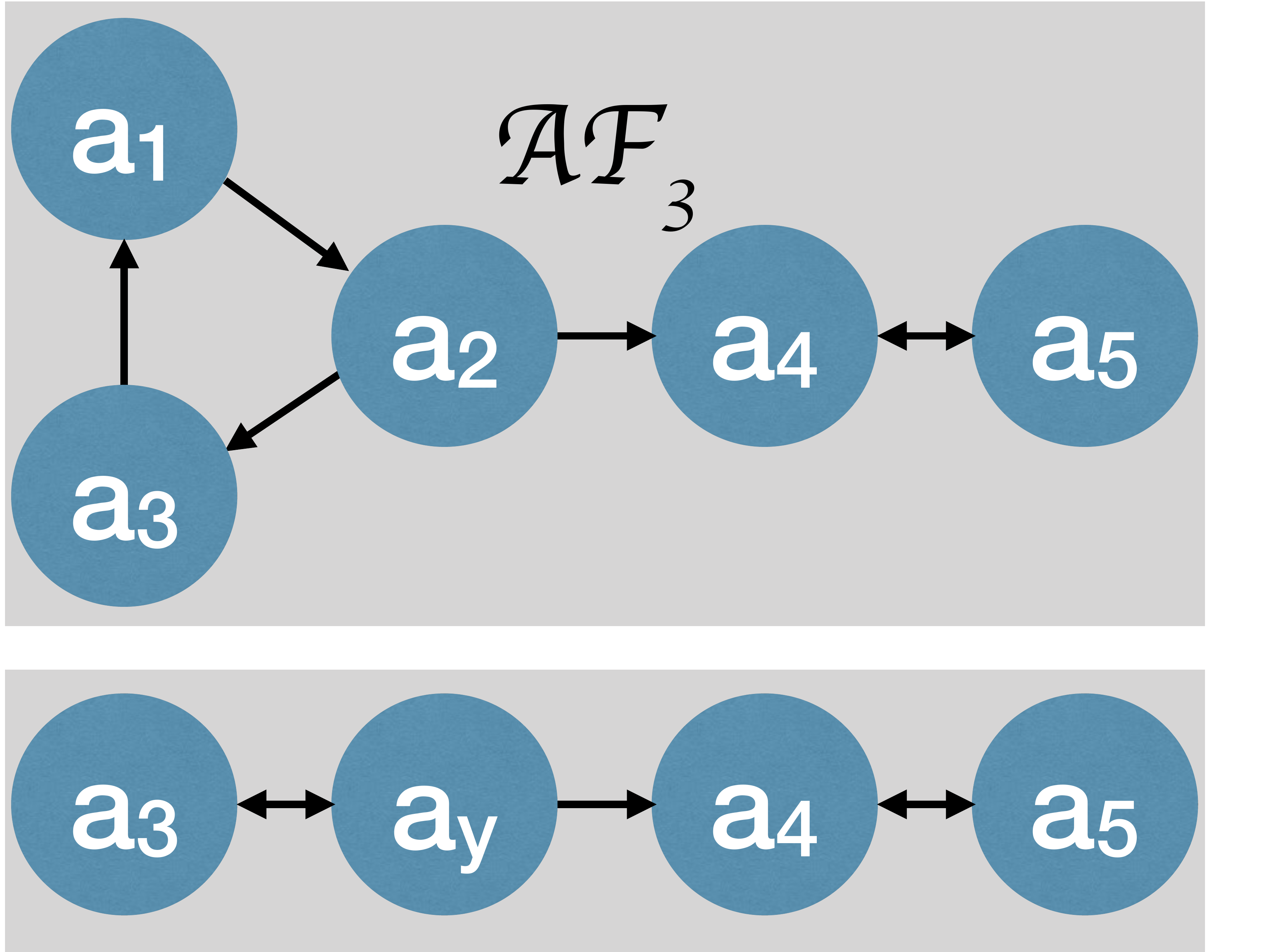}  
      \includegraphics[scale=0.11]{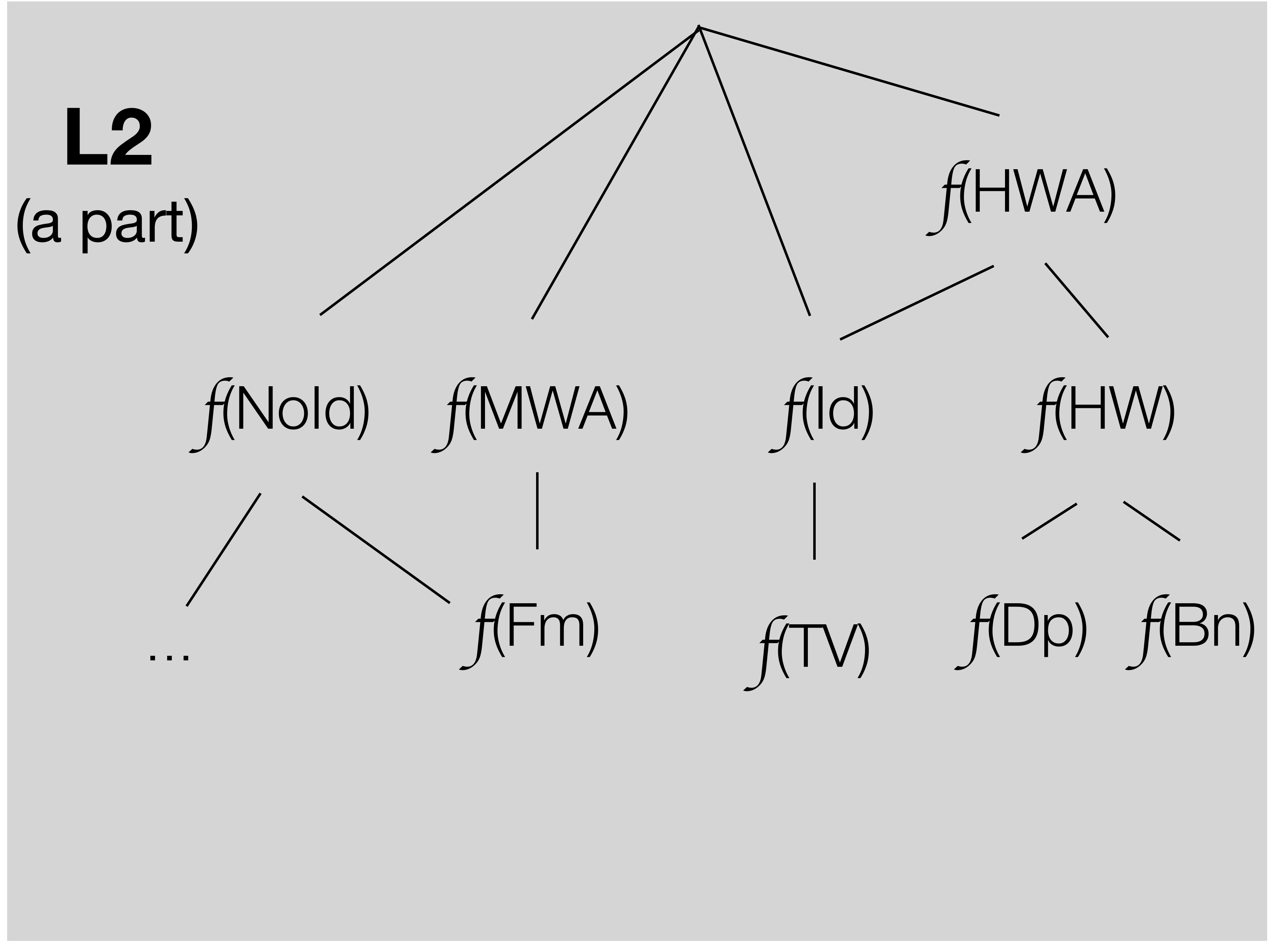}
    \end{center}  
    \caption{\normalfont Top left: an argumentation framework $AF_3$. 
     Bottom-left: $AF_3$'s abstraction. Right: 
     an abstract lattice $L_2$.} 
    \label{marathon} 
\end{figure}

We conclude this section with comparisons to 
Dung preferred semantics and cf2 semantics \cite{Baroni05}.  
Let us first consider $AF_1$ in Figure A and 
the lattices as shown in Figure D.  
Let us denote $g_\gamma^{AF} \circ g_p \circ g_{\alpha}$ by $\mathbb{G}^{AF}$,
then we have: $\emptyset$ for $g_p(AF_1)$ (i.e. Dung preferred set); 
$\{\{a_1, a_5\}, \{a_2, a_5\}, \{a_3, a_5\}\}$ for cf2($AF_1$); 
while $\{a_5\}$ for $\mathbb{G}^{AF_1}(AF_1)$ (as we have already
shown the only one abstract space argumentation framework, 
in Figure B, we omit the derivation process). 
By comparisons between $g_p(AF_1)$ and $\mathbb{G}^{AF_1}(AF_1)$, 
we observe that all $a_1,a_2, a_3, a_4$ are {\it -approved}, while 
$a_5$ is {\it implied}. Hence in this case, with respect to 
the semantic structure of $L_2$, we might say that 
Dung preferred set behaves more conservative than necessary. 
On the other hand, by comparisons between cf2$(AF_1)$ and $\mathbb{G}^{AF_1}(AF_1)$, 
we observe that cf2$(AF_1)$ accepts either of the arguments in the odd 
cycle, 
which is more liberal than necessary with respect to 
$L_2$ - since no arguments in $AF_1$ could break
the preference 
pre-order $\focusOnOs < \focusOnMp < \focusOnEc < \focusOnOs$ 
of the three arguments. 
Therefore, for $AF_1$, Dung semantics seems to give false-negative 
to $a_5$ acceptability, while cf2 seems to give false-positives 
to either of $a_1$, $a_2$, $a_3$ acceptability. If 
those acceptability semantics aim to answer ``Which arguments 
should be (credulously) accepted?", false-negatives only signal 
omission, but false-positives signal unintuitive results and are less 
desirable. \\ 
%\indent Let us secondly consider $AF_2$ in Figure C.  
%In this example, we have: $\{\{a_6, a_9\}, \{a_7, a_8\}\}$ for both $g_p(AF_2)$ 
%and cf2$(AF_2)$, while $\emptyset$ for $\mathbb{G}^{AF_2}(AF_2)$. 
%All the arguments of $AF_2$ are {\it questioned} here. 
%This example illustrates that there are situations 
%where neither Dung preferred semantics, nor cf2, generalisation of Dung 
%preferred semantics, 
%offers a good prediction. \\
\indent Let us, however, consider another argumentation framework 
$AF_3$ in Figure \ref{marathon} borrowed from 
\cite{Baroni05}. 
Consider: 
\begin{description} 
    \item[$a_1$] The downpour has been relentless since the morning.  
    \item[$a_2$] It was burning hot today. 
    \item[$a_3$] All our employees ran a pleasant full marathon today.
    \item[$a_4$] Nobody stayed indoor. 
    \item[$a_5$] Many enjoyed TV shows at home. 
\end{description}   
%Consider the following arguments: $a_1$: The downpour has been relentless since the morning; $a_2$: It was burning hot today; $a_3$: All our 100 employees ran a pleasant full marathon today; $a_4$: Nobody stayed indoor; 
%	$a_5$: Many enjoyed TV shows at home.
We assume the abstract lattice as shown in Figure \ref{marathon} 
for $AF_3$. We assume $M = \bigvee L_2$, and any 
nodes below $f(\text{Fm})$, $f(\text{TV})$, $f(\text{Dp})$, $f(\text{Bn})$
not explicitly 
shown there are still assumed to be there. W, A, M, H, Id, 
Fm, Dp, Br each abbreviates Weather, Activity, 
Mild, Hard, Indoor, Full-marathon, Downpour and Burning. 
The lattice expresses in particular that a downpour and  
the burning sun relate under the hard weather, and 
the hard weather and indoor activities such as watching TV shows 
relate under hard weather activity (that is, an activity to do  
under a hard weather condition), but that  
hard weather and mild weather activities do not go together. Also, 
indoor and no-indoor oppose. Here we have: $\{\{a_5\}\}$ for $g_p(AF_3)$; $\{\{a_1, a_4\}, \{a_1, a_5\}, 
\{a_2, a_5\}, \{a_3, a_4\}, \{a_3, a_5\}\}$ for cf2$(AF_3)$. Meanwhile,
for $\mathbb{G}^{AF_3}(AF_3)$, $\{a_1, a_2\}$ is first of all 
the set of a maximal subset of $\{a_1, a_2, a_3\}$. 
It is attack-preserving: $\alpha(\{\text{DP}, \text{Br}\}) = f(\text{HW})$,
which is not comparable with $f(\text{NoId})$ or $f(\text{Fm})$, 
valid because $f(\text{HW})$ does not abstract Fm, non-trivial, 
and compatible. Hence the argumentation framework shown 
under $AF_3$ in Figure \ref{marathon} is 
the abstract space argumentation framework with respect to $L_2$. 
Consequently, 
$\mathbb{G}^{AF_3}(AF_3) = \{\{a_3, a_4\}, \{a_3, a_5\}, \{a_5\}\}$. 
Therefore, in this example, $\mathbb{G}^{AF_3}(AF_3)$, 
too, credulously accepts an argument in the odd-cycle as cf2($AF_3$) does. 
Notice, however, 
that we still obtain the Dung conservative preferred set $\{a_5\}$ which 
obtains from $\{a_y, a_5\}$. \\
\indent It is safe to observe that 
the traditional Dung, or cf2, which is more appropriate depends 
not just on an argument graph but also the semantic relation among 
the arguments in the graph; and that combination 
of abstract argumentation and abstract interpretation is one viable 
methodology to address this problem around cycles in argumentation 
frameworks.

\section{Related work}    

As far as we are aware, this is the first study that incorporates abstract 
interpretation into abstract argumentation theory. %and - as we so suppose - 
%also the first 
%study that genuinely treats all cycles, regardless of the cycle length, 
%uniformly, in trying to address certain anomaly that has pertained 
%to the odd-ness or even-ness of the length of a cycle in a 
%Dung argumentation framework. \\
%\indent 
Odd-sized cycles 
have been a popular topic of research in the literature for some time, 
as they tend to prevent the acceptability of all subsequent 
arguments with respect to directionality. 
Noting the difference between preferred and the grounded semantics, 
Baroni et al. \cite{Baroni05} proposed to 
accept maximal 
conflict-free subsets of a cycle 
for gaining more acceptable arguments off an odd-length 
cycle, which led to cf1/cf2 semantics. They 
are regarded as improvements on more traditional naive semantics 
\cite{Bondarenko97}. They also weaken Dung defence around strongly 
connected components of an argumentation framework 
into SCC-recursiveness. \\
\indent The stage2 semantics that took inspiration 
from cf2 is another approach with a similar SCC-recursive aspect, but 
which is based on the stage semantics \cite{Verheij96} rather than 
the naive semantics, the incentive being to maximise 
range (the range of a set of arguments is itself plus all arguments 
it attacks). \\
\indent The fundamental motivation behind those semantics was to treat 
an odd-length cycle in a similar manner to an even-length cycle. 
As we showed, however, specialisation of Dung semantics 
without regard to semantic relation among arguments in a given argumentation 
framework seems not fully generalisable. 
%there are even-length cycles 
%that ought to be treated as odd-length cycles (see the second 
%example in Introduction). In fact, some odd-length cycles 
%and some even-length cycles in a Dung argumentation graph 
%should really behave as how even-length and respectively odd-length 
%cycles behave in a Dung argumentation graph, while some odd-length cycles
%and some even-length cycles should be treated 
%indeed the same way Dung treats odd-length and respectively even-length 
%cycles. The problem
%is that it is almost impossible to know purely argumentgraph-theoretically 
%which category an 
%odd or an even cycle  would belong to, 
%for we can always come up with a particular instance of 
%$AF_1$ for which not accepting any of its arguments appears to be the only 
%one sensible 
%solution, or of $AF_2$ for which accepting either 
%$\{a_6, a_9\}$ or $\{a_7, a_8\}$ seems to be the only sensible. \\
\indent To an extent, that any such systematic resolution of 
acceptability of cyclic arguments based only on a Dung 
argumentation graph is tricky relates
to the fact that attacking arguments  
in a cycle can be contrarily \cite{Horn89} but not necessarily 
contradictorily 
opposing.
As the study in \cite{Baroni15} shows and as is known in linguistics, 
dealing with contrary relations is difficult in Fregean logic. 
However,  with abstract interpretation, we can take advantage 
of semantic information of arguments in partitioning 
those attacking arguments in a cycle into mutually incompatible 
subsets, by which uniform treatment of cycles come into reach.

\section{Conclusion} 

We introduced abstract interpretation into argumentation frameworks. 
Our formulation shows it is also 
a powerful methodology in AI reasoning. 
We believe 
that more and more attention will 
be directed towards semantic-argumentgraph hybrid 
studies within argumentation community, 
and we hope that our work will provide one fruitful research direction.  

\section*{Acknowledgement} 
We thank Leon van der Torre
and Ken Satoh for  discussion on related topics which greatly influeced this work
of ours.

\bibliographystyle{aaai}
\bibliography{references} 
\end{document}